\documentclass[11pt]{article}
\usepackage[T1]{fontenc}
\usepackage[utf8]{inputenc}
\usepackage{verbatim}
\usepackage{booktabs}
\usepackage{mathtools}
\usepackage{amsmath}
\usepackage{amsthm}
\usepackage{amssymb}
\usepackage{graphicx}
\usepackage{setspace}
\usepackage{cleveref}

\usepackage[authoryear]{natbib}
\PassOptionsToPackage{normalem}{ulem}
\usepackage{ulem}
\makeatletter

\providecommand{\tabularnewline}{\\}

\theoremstyle{plain}
\newtheorem{prop}{\protect\propositionname}
\theoremstyle{remark}
\newtheorem{rem}{\protect\remarkname}

\@ifundefined{date}{}{\date{}}

	\newcommand{\blind}{0}
    \renewcommand{\section}{\@startsection {section}{1}{\z@}%
                                       {-3.5ex \@plus -1ex \@minus -.2ex}%
                                       {2.3ex \@plus.2ex}%
                                       {\normalfont\fontfamily{phv}\fontsize{16}{19}\bfseries}}
    \renewcommand{\subsection}{\@startsection{subsection}{2}{\z@}%
                                         {-3.25ex\@plus -1ex \@minus -.2ex}%
                                         {1.5ex \@plus .2ex}%
                                         {\normalfont\fontfamily{phv}\fontsize{14}{17}\bfseries}}
    \renewcommand{\subsubsection}{\@startsection{subsubsection}{3}{\z@}%
                                        {-3.25ex\@plus -1ex \@minus -.2ex}%
                                         {1.5ex \@plus .2ex}%
                                         {\normalfont\normalsize\fontfamily{phv}\fontsize{14}{17}\selectfont}}
    
\usepackage{amsthm}
\usepackage{amsfonts}

\usepackage{enumerate}

\usepackage{bm}

\usepackage{soul}
\usepackage{caption}
\usepackage{subcaption}
\usepackage{float}
\usepackage{todonotes}
\DeclareMathOperator*{\argmax}{arg\,max}

\global\long\def\g{\,|\,}%
\global\long\def\mbx{\bm{x}}%
\global\long\def\mbtheta{\bm{\theta}}%
\makeatother

\providecommand{\propositionname}{Proposition}
\providecommand{\remarkname}{Remark}

\begin{document}
 
\global\long\def\spacingset#1{
\global\long\def\baselinestretch{%
}%
\small\normalsize}%
 \spacingset{1} 

\title{\textbf{Image-based Novel Fault Detection with Deep Learning Classifiers
using Hierarchical Labels}}
\author{Nurettin Sergin$^{a}$, Jiayu Huang$^{a}$, Tzyy-Shuh Chang$^{b}$,
Hao Yan$^{a}$ \\
 $^{a}$School of Computing and Augmented Intelligence, Arizona State
University, Tempe, USA \\
 $^{b}$OG Technology, Ann Arbor, MI, USA }

\maketitle

\if1\blind {

\author{Author information is purposely removed for double-blind review}

\bigskip{}
\bigskip{}
\bigskip{}

\begin{center}
\textbf{\LARGE{}Image-based Novel Fault Detection with Deep Learning
Classifiers Using Hierarchical Labels}
\par\end{center}

\medskip{}
} \fi \bigskip{}

\begin{abstract}
One important characteristic of modern fault classification systems
is the ability to flag the system when faced with previously unseen
fault types. This work considers the unknown fault detection capabilities
of deep neural network-based fault classifiers. Specifically, we propose
a methodology on how, when available, labels regarding the fault taxonomy
can be used to increase unknown fault detection performance without
sacrificing model performance. To achieve this, we propose to utilize
soft label techniques to improve the state-of-the-art deep novel fault
detection techniques during the training process and novel hierarchically
consistent detection statistics for online novel fault detection.
Finally, we demonstrated increased detection performance on novel
fault detection in inspection images from the hot steel rolling process,
with results well replicated across multiple scenarios and baseline
detection methods. 
\end{abstract}

\spacingset{1.5} 

\section{Introduction}

Many manufacturing systems are instrumented with image-sensing systems
to monitor process performance and product quality. The low cost and
rich information of the image-based sensing systems have led to high-dimensional
data streams that provide distinctive opportunities for performance
improvement. Among these, accurate process monitoring and fault classification
are among the benefits gained from the rich information these image
sensors can provide. In literature, process monitoring often refers
to the step of detecting and isolating abnormal samples in a certain
process. Normally, after process monitoring, fault classification
is performed, and the isolated fault is classified into one or more
known types of fault. Fault classification is an essential step within
the process monitoring loop, at which point the type of detected and
identified faults are determined \citep{Chiang2001-nu}. Accurate
fault classification can provide engineers with favorable information
to isolate and diagnose system faults and anomalies to improve quality
and maximize system efficiency.

However, fault classification in manufacturing systems typically assumes
a fixed set of fault modes. In this case, the existing fault classification
model may make overconfident decisions or fail silently and, at certain
times, dangerously for new unseen fault types. An additional policy
should be developed to flag the emergence of unseen faults to alert
the system and log the related instances for human evaluation and
subsequent updates of the model. Such practice has been implemented
in some monitoring systems in modern machine learning models \citep{BreckCNSS17}.
We name this problem image-based novel fault detection, which will
be the major focus of this paper.

The recent resurgence in machine learning research heavily influenced
anomaly detection, novel fault detection, and fault classification
literature \citep{Liu201833}. For example, feature-based and tensor-based
anomaly detection and fault classification methods have been proposed
\citep{yan2014image}. However, the traditional methods typically
rely on the accurate definition of the handcrafted features, which
often require much labor work and also lead to unsatisfying performance.
Deep learning, a subclass of machine learning methods, has attracted
a peculiar interest. What makes deep learning methods so attractive
for anomaly detection \citep{kwon2019survey,sergin2021toward} or
fault classification \citep{pan2017liftingnet} is their ability to
learn useful feature representation for data representation or class
discrimination automatically from data as opposed to requiring handcrafted
ones, which is often challenging in fault classification. Despite
the impressive performance increases, deep learning has brought the
the problem of fault classification and anomaly detection. These models
either assume that no labels are available (e.g., anomaly detection)
or assume a closed set of fault types by using a fixed number of neurons
in the last layers (e.g., classification), which cannot be used in
novel fault detection.

To address the novel fault detection problem, some literature has
recently proposed out-of-distribution methods to detect new emerging
classes. In these works, a set of fault classes is assumed to be
given, and the goal is to detect novel or unseen faults in the system.
For more details about the novel fault classification, we will provide
a complete literature review in \Cref{subsec:ood-detection}.

\begin{figure*}[!t]
\centering \includegraphics[width=0.9\linewidth]{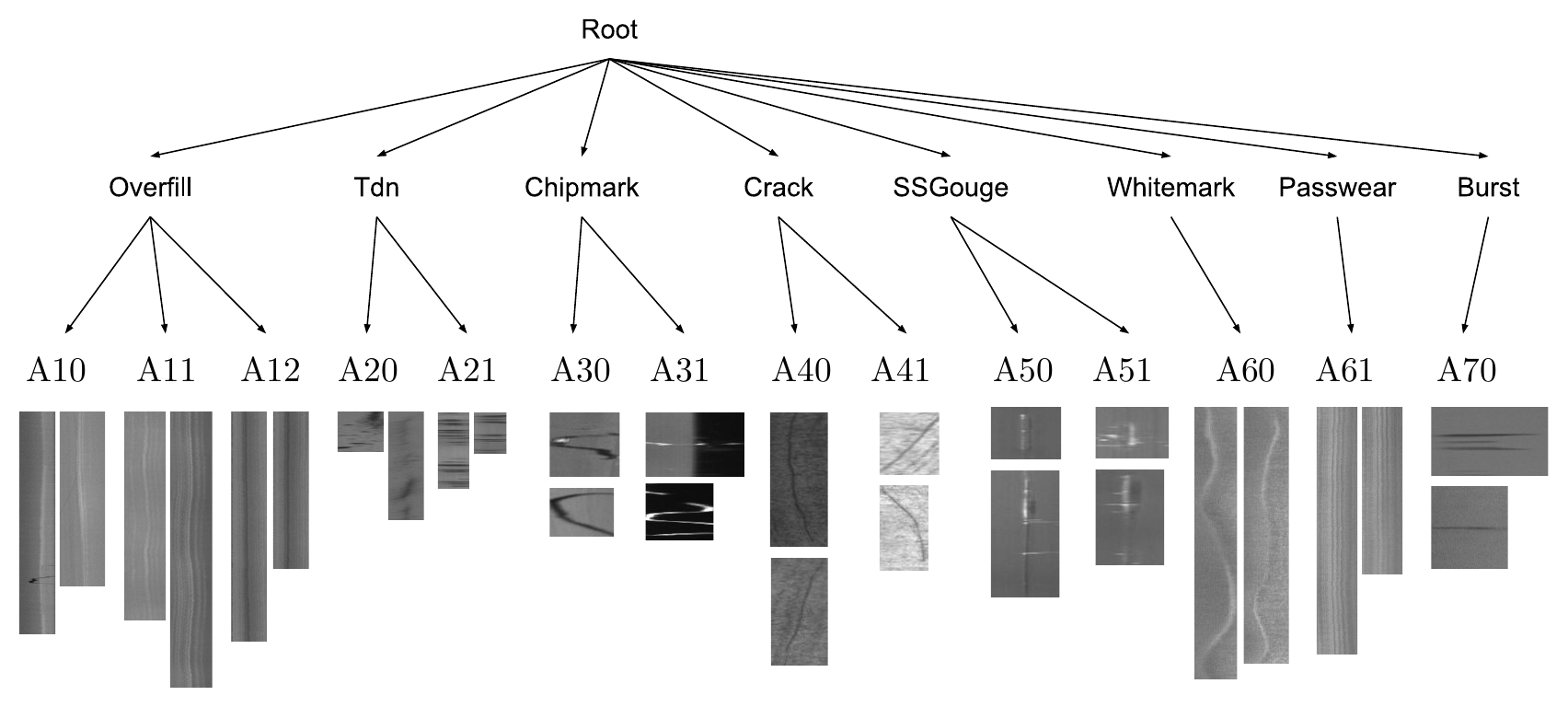}
\caption{Graphical illustration of the hierarchy of defects in the
hot steel rolling dataset.}
\label{fig:graphical-illustration-hierarchy-rolling} 
\end{figure*}

However, in many existing complex systems, the classes typically can
be represented in a hierarchical tree structure rather than a flat
structure. 
In modern manufacturing systems, many fault classes exist and the
fault classes can often be represented in a hierarchical structure.
For example, for the defect inspection problem in the rolling manufacturing
system, 14 different types of anomalies have been identified. These
14 anomalies can be classified into 8 subcategories, given the shapes
of the defect areas. Here, many fault classes may have a very small
sample size \citep{sahoo2003critical}. Such a hierarchical fault
structure is common in manufacturing systems due to the hierarchical
root causes contributing to product quality often represented by the
fishbone diagram. The hierarchical set of factors organized by the
fishbone diagram may also lead to a similar set of hierarchical fault
patterns. An example of such a hierarchical fault pattern in a steel
rolling process inspection is shown in \Cref{fig:graphical-illustration-hierarchy-rolling}.
The first level determines which major category of defect this inspection
image patch belongs to. Some examples of these defect categories are
chip marks, cracks, and overfill. The second level defines the subcategory
within each major category. For example, the overfilled major category
contains black (A12), white (A11), and lite (A10) subcategories. The
major shape of these categories are similar (showing verticle patterns),
but the color is different for each subcategory.

In the literature, there are many existing works on hierarchical fault
classification methods considering hierarchical fault structures.
In these works, hierarchical classification methods have demonstrated
improved classification accuracy over common flat classifiers. The
improvement in performance is mostly due to the following two major
reasons: First, it can guide the learning process due to the injected
inductive bias and result in a more accurate estimation. Second, the
knowledge learned from the other fault nodes can be transferred to
the rare fault modes by utilizing the relationship between the fault
modes within the same ancestor. Finally, the incorporation of the
hierarchical information can help understand the severity of the mistakes
in the training process, therefore improving the overall classification
accuracy. The motivation behind this is that if the classifier fails
to identify the right class, it may still predict it as a known class
that is closer to the ground truth in the hierarchy (i.e., share the
same parent node).

However, even though a large number of works for hierarchical classification
exist, to the best of the authors' knowledge, there is currently no
deep learning-based novel fault detection methods considering this
hierarchical fault taxonomy.

In this work, we propose a novel approach such that the knowledge
of the fault hierarchical taxonomy can be incorporated into both the
training procedure and the online novel fault detection procedure
by combining soft labeling and hierarchically consistent score in
deep learning. We demonstrate how a model trained in consideration
of the hierarchical relationship between the labels will be more effective
at detecting previously unseen fault classes. We further demonstrate
that the proposed method can improve upon state-of-the-art deep learning-based
fault detection methods. While many fault sets are inherently hierarchical
(e.g., fault trees), oftentimes this information is neglected in favor
of the simplicity of using only the label information at leaf nodes.
With this work, we aim to convince the practitioners to inject this
information into the training procedure and the detection statistics
for anomaly detection.

The main contribution of this paper can be summarized as
follows: 
\begin{itemize}
\item We propose a novel methodology by considering the hierarchical structure
of taxonomy in real-world novel fault detection by incorporating it
into a deep-learning model classifier. By injecting hierarchical information
into the loss function in the deep learning model, we consider both
classification accuracy and hierarchical structure consistency during
model training. 
\item We propose a hierarchically consistent score to detect unknown faults
online. The primary purpose of this score is to identify samples that
are inconsistent with the known fault hierarchical taxonomy, improving
the overall performance of the classifier-based out-of-distribution
detector. 
\item We provide some methodology insights into why including such hierarchical
information in offline training and introducing hierarchical consistency
scores for online detection can improve the novel fault detection
accuracy. Additionally, we propose two propositions to illustrate
the insights of our proposed score and visualization of why hierarchical
training would be beneficial. Finally, we discuss the practical implementation
of the proposed methods for a rolling process with hierarchical structures
on the anomaly taxonomy and demonstrate the improved accuracy for
novel fault detection. 
\end{itemize}

The rest of the article is organized as follows. In \Cref{sec:background},
we provide a background of related works in the out-of-distribution
detection and hierarchical classification literature. In \Cref{sec:methodology},
we provide the details of the interventions that we propose at training
and test time. In \Cref{sec: case study}, we introduce our hot
steel rolling image defect dataset, formalize our experiment design
and define the performance evaluation criteria we consider. The results
of the experiments and related discussions are presented in \Cref{sec:results}.
Finally, we conclude the work in \Cref{sec:conclusion}, and a brief
data availability statement will also be included in \Cref{sec:dataavail}.

\section{Related Work \label{sec:background}}

This section briefly reviews the literature related to out-of-distribution
detection in \Cref{subsec:ood-detection} and the use of hierarchical
classification to address the hierarchical labeling relationship in
\Cref{subsec:hierarchical-classification}, as these two methods
are the fundamental pillars of the proposed fault detection algorithm
that is discussed in the next section.

\subsection{Out-of-distribution Detection \label{subsec:ood-detection}}

As mentioned in \citep{BendaleB16}, deep learning classifiers lack
the ability to detect unknown faults due to the implicit closed-set
assumption, which states that all classes are known as priori \citep{BendaleB16}.
This closed-set assumption prevents the deep-learning methodology
from applying to safety-related applications. The research on out-of-distribution
(OOD) detection focuses on detecting anything other than
what lies in the training set.

The definition of what can be considered out-of-distribution is broad.
Samples from previously unseen classes, adversarial examples \citep{GoodfellowSS14}
or domain shift \citep{Ben-DavidBCKPV10} can all be considered 
OOD.  

\subsubsection{Classifier-based models}

The literature on OOD detection reveals two primary
approaches: 1) classifier-based models, which differentiate between
normal and anomaly based on a specific optimization function. 2) generative
models, which involve modeling the data's distribution and detecting
anomalies based on deviations from learned distribution or reconstruction,
such as the Variational Auto-encoder (VAE) and Generative Adversarial
Networks (GAN).  %

In the out-of-distribution detection combined with the classification
problem, the uncertainty of the labels, called aleatoric uncertainty
\citep{kendall2017uncertainties}\citep{sallak2013reliability}, is deserved to be considered since
the samples we obtained could be a novel and unseen class and the
neural network outputs are often probabilities. This type of uncertainty
can be used as an indicator of anomaly samples. The rest of the section
will demonstrate the aleatoric uncertainty usage for out-of-distribution
detection in some methods.

Two main types of classifier-based methods have been developed.
The first class focuses on utilizing Maximum Softmax Probability (MSP)
\citep{HendrycksG17}. MSP is the largest probability value produced
by the deep learning classifiers. Their major observation in \citep{HendrycksG17}
was that probability assigned to the most likely class (i.e., MSP)
differed considerably between in-distribution and out-of-distribution
samples. For example, the probability distribution of the out-of-distribution
samples tend to be closer to the uniform distribution compared to
the out-of-distribution samples. In the MSP method, the idea of the aleatoric
uncertainty difference between in-distribution and out-of-distribution
samples is applied. Later, many efforts have been contributed to improving
the use of MSP for out-of-distribution detection. Among those methods,
the Out-of-DIstribution detector for Neural networks (ODIN) \citep{LiangLS18}
has improved the performance of MSP by modifying the training regime
with temperature scaling and adversarial perturbation, which significantly
improves the separation between in-distribution and out-of-distribution
samples. According to the review paper \citep{shafaei2018less}, ODIN
has achieved state-of-art performance in all MSP-based methods.

The second line of work focuses on the neuron response values at different
levels to detect samples from novel classes. This method is originally
proposed by \citet{LeeLLS18}, where the authors observe that neuron
response outputs at various levels of a neural network differ significantly
from one class to the other, as well as the in-distribution class to
the out-of-distribution class. Therefore, the neuron response of a
new sample can be compared with the distribution of the response from
the existing known classes, where the sample-to-distribution distance
can be used as an abnormality score. For example, multivariate Gaussian
distributions with tied covariances at each level are used for modeling
the neuron response for each known class, and the Mahalanobis distance
is used to define a potential new class in the samples. While experiments
suggest further improvements in accuracy, this method relies on additional
out-of-distribution datasets for the tuning of the hyper-parameters.

We also would like to mention that there are some other
frameworks besides these two lines of research by utilizing the out-of-distribution
samples to improve the out-of-distribution detection performance.
Among these methods, \citet{HendrycksMD19} proposed a complementary
method to the existing scoring-based out-of-distribution detection
methods. The method, named Outlier Exposure, simply fine-tunes an
already trained model, again with the use of the auxiliary outlying
dataset, to encourage even more separation between in-distribution
and out-of-distribution, regardless of the scoring method used. However,
these methods assume that the out-of-distribution samples are available,
which is not applicable in the application of novel fault detection.

\subsubsection{Generative models }

Generative models such as VAE and GAN have been widely used for anomaly
detection via reconstruction error. However, their performance in
detecting out-of-distribution (OOD) data has been questioned, particularly
for image data, given their focus on detecting global patterns rather
than local anomalies \citep{nalisnick2018deep,havtorn2021hierarchical}.
Additionally, VAE methods have been criticized for generating blurry
images with little detail, limiting their effectiveness for OOD detection.
While VAE-based OOD detection techniques have been developed, they
have primarily been evaluated on natural image datasets with very
clear intra- and inter-class variations, such as MNIST, FashionMNIST,
.etc \citep{guo2021cvad}. GAN methods have been criticized for being
prone to mode collapse \citep{zhang2021mode,murray2021performance},
where they only generate a subset of the possible variations of the
target distribution, leading to a limited representation of the data.

Some initial works have been applied to OOD for industrial image inspection
\citep{sergin2021toward,yan2019image}. However, in many classification-based
OOD problems, where the class information is available, we are interested
in detecting the unseen anomaly, given that multiple classes exist.
Traditional VAE or GAN lacks the ability to encode the class information.

Even though in literature, there are some existing works on encoding
the class information into generative models, such as conditional
VAE \citep{mirza2014conditional}, Knowledge-oriented VAE \citep{shen2019kmr}
and conditional GAN \citep{sohn2015learning}. However, in literature,
the class information is often represented as the flat structures,
instead of the hierarchical structure used in this paper.

Several comparison studies in the literature have shown that generative
models generally have poorer performance than most other classifier-based
methods, such as ODIN, MSP, and DMD, especially when class information
is available \citep{li2022deep}. Furthermore, training a generative
model such as VAE or GAN is typically more expensive than training
a classifier, making them less practical for practitioners.

To the best of our knowledge, there has been no exploration of incorporating
hierarchical label structure into deep learning models for improving
the performance of existing classifiers or other generative methods
in OOD detection.

In this paper, we propose a method that leverages the statistical
properties of deep learning classifiers by injecting hierarchical
knowledge into their optimization functions. The proposed hierarchical
models have the potential to perform better at detecting unseen anomalies.

\subsection{Hierarchical Classification}

\label{subsec:hierarchical-classification}

In this subsection, we will first make a clear definition of hierarchical
classification. We will then introduce the existing literature as
well as the advantages of a hierarchical classifier over a flat classifier.

First, a formal definition of the hierarchical classification model,
originally proposed by \citet{SillaFreitasSurveyOfHC11}, can be given
with the following three assumptions: 
\begin{enumerate}
\item A hierarchy can be modeled either by a tree or a directed acyclic
graph, depicting subsumption relationships among classes. Each class
is subsumed under its parent(s) in this structure. 
\item The structure is pre-defined and provided by the modeler. It is not
meant to be learned from the data. 
\item Each data point belongs to one and only one of the leaf nodes in the
hierarchy. 
\end{enumerate}
Given this structure, a simple approach to classification would be
flattening out all the leaf nodes and treating the classification problem as an $N$ -way flat classification task where there are,
in total, $N$ leaf nodes. However, this structure treats each class
as equally different from the other and completely ignores the rich
information provided by the hierarchy.

In literature, many existing methods utilize hierarchical classification
to improve the classification accuracy of different tasks, such as
bioinformatics \citep{freitas2007tutorial}, community data \citep{gauch1981hierarchical},
web content \citep{dumais2000hierarchical}, accident report \citep{zhao2021hierarchical},
e-commerce \citep{ShenLSICforEcommerce12}, and visual recognition
\citep{YanHDCNN15}. Overall, \citet{silla2011survey} divided the
existing approaches into the top-down local classifier approach and
the global-classifier approach.

The top-down local classifier is originally proposed by \citet{koller1997hierarchically}.
More specifically, these approaches can be further divided into a
local classifier per node \citep{fagni2007selection}, a local classifier
per parent node \citep{brown2012experimental}, and a local classifier
per level \citep{clare2003predicting}, depending on how the local
information is considered. However, one of the major disadvantages
of the top-down approach is that any error at a certain class level
will be propagated down the hierarchy. Please see \citep{silla2011survey}
for a more detailed review of the local classifier approach.

To overcome the drawback of the local classifier approach, global
classifiers are proposed by \citet{xiao2007hierarchical}, which aims
to consider a single classification model trained from the training
set, taking into consideration of the class hierarchy as a whole during
a single run of the classification algorithm. Typically, learning
a single model for all classes has the advantage that the size of
the global classification model is considerably smaller, which is
especially important for deep-learning-based models, given that deep-learning models are already large. Moreover, the dependencies between
the classes can be taken into account in a more natural and straightforward
way. Currently, many approaches exist, such as multi-label classification
methods \citep{kiritchenko2005functional} and distance-based methods
\citep{rocchio1971smart}. The multi-label classification approach represents
each label and its structure using multiple attributes, and the algorithm
aims to predict all attributes simultaneously. Distance-based methods
aim to assign the new sample to the nearest class by computing the
distance between the new test example and each class, considering
the hierarchical structure in the distance measure. Recently, a new
global hierarchical classification model has been proposed \citep{bertinetto2020making}
by defining the soft labels to represent the hierarchical structure
of the class. Such a method is simple to implement while achieving
state-of-the-art performance in the ImageNet classification. Therefore,
this paper focuses on incorporating the ``soft-label'' approach
toward the OOD performance with the hierarchical label
structure.

\section{Proposed Methodology \label{sec:methodology}}

We present the mathematical framework related to the key elements
of the neural network-based novel fault detection systems that will
be used throughout this section.

Assume we have a dataset $\mathcal{D}=\{(\mbx^{n},y^{n})\g n\in\{1\dots N\}\}$
where $\mbx^{n}$ are inputs and $y^{n}\in\{1\dots K\}$ denote which
of the $K$ labels that $\mbx^{n}$ belong to. In the context of fault
classification, $\mbx$ can be various inputs, such as images collected
from automated optical inspection cameras or readings from sensors.
Each input is associated with one and only one fault label, $y$,
which is assumed to be drawn from a categorical distribution with
$K$ distinct fault types. We can assume that this dataset is generated
by a random process governed by the joint distribution $p(\mbx,y)$.

The aim of a neural network-based fault classifier is to approximate
the conditional distribution $p(y\g\mbx)$ with a function $g(\mbx,\mbtheta)$
parameterized by neural network weights $\mbtheta$ such that the
class with the highest predicted probability $\hat{y}$ will be assigned
to the input $\mbx$. The function $g(\mbx,\mbtheta)$ represents
the penultimate layer of the neural network. Here, the penultimate
layer often refers to the feature layer before the Softmax layer.
A common practice is to project this representation into the probability
simplex by using a Softmax layer. We shall denote this function whose
codomain is the probability simplex as $f(\mbx,\mbtheta)$. To simplify
notation, $f(\mbx)$ and $g(\mbx)$ will be used instead of $f(\mbx,\mbtheta)$
and $g(\mbx,\mbtheta)$ respectively, which implies that the model
has been trained on the training data and the parameters are fixed.
The output of the penultimate layer for samples in class $k$ is denoted
by the subscript, $g_{k}(\mbx)$, and the respective probability prediction
is $f_{k}(\mbx)$. In another word, we have $f_{k}(\mbx)=\mathrm{Softmax}(g_{k}(\mbx))$.

\begin{figure}[!t]
\centering \includegraphics[width=0.85\linewidth]{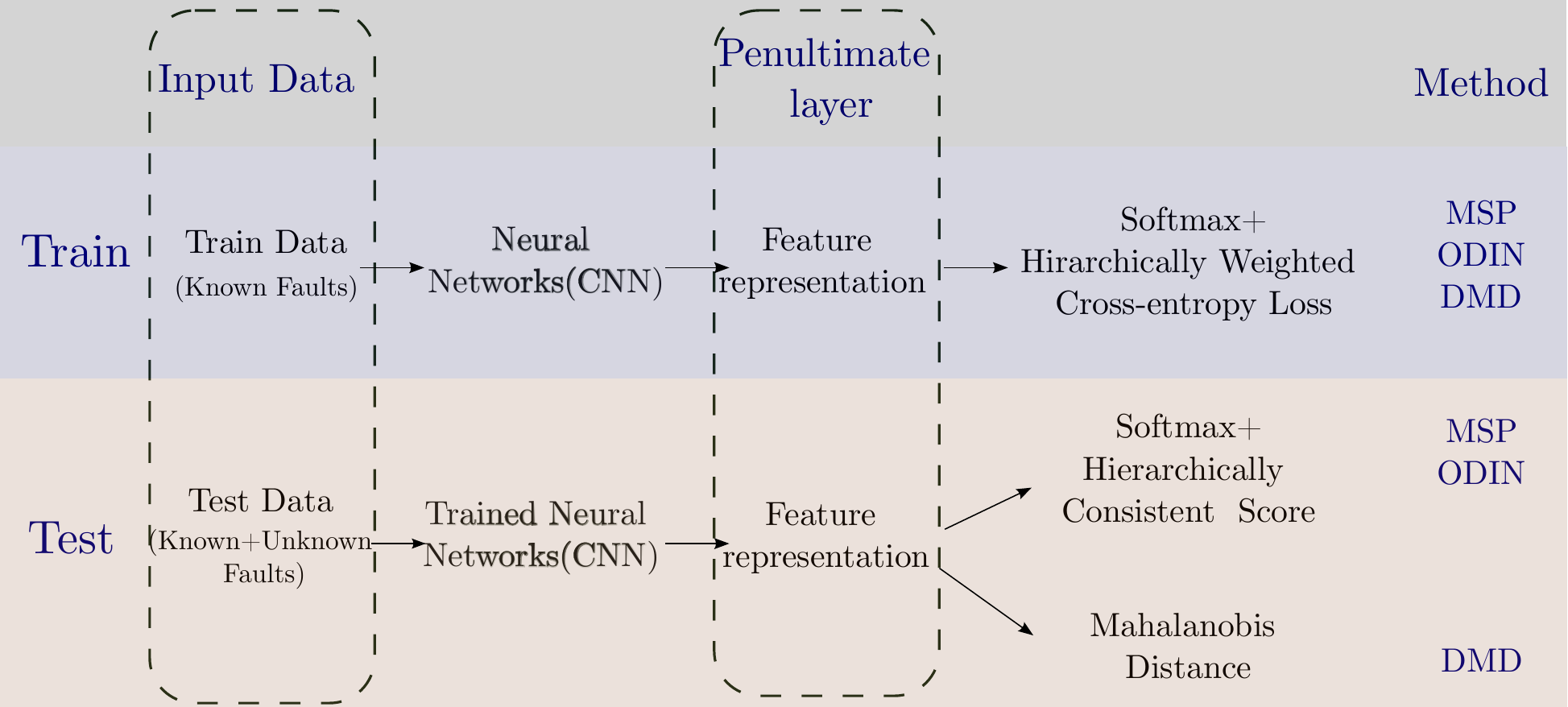}
\caption{Methodology Flowchart}
\label{fig:methodology} 
\end{figure}

This section will start with the review of three baseline methods
in \Cref{subsec: OODbasline}, including the Maximum Softmax Probability
(MSP), Out-of-Distribution detector (ODIN), and Deep Mahalanobis Detection
(DMD). In \Cref{subsec:hierarchical-reg-anom-score}, we demonstrate
that in the training stage, hierarchical regularization via soft labeling
modification in the loss function will be incorporated. In \Cref{subsec:metho},
a novel hierarchically consistent score function for novel fault detection
will be illustrated for three baseline methods. \Cref{fig:methodology}
illustrates a summary of the methodology in this section.

\subsection{Formulation of baseline OOD detection scoring methods used in this
study \label{subsec: OODbasline}}

In this subsection, we will review three state-of-the-art OOD classifiers
in three subsections, namely the Maximum Softmax Probability (MSP),
Out-of-Distribution detector (ODIN), and Deep Mahalanobis Detection
(DMD).

\subsubsection{Maximum Softmax Probability}

Maximum Softmax probability (MSP) was introduced in \citep{HendrycksG17}.
The idea of MSP is intuitive, where the in-distribution samples turned
out to have a smaller uncertainty or a larger prediction probability for
a certain class. On the other hand, out-of-distribution samples may
have a much more uniform distribution for all the known classes. Here,
the maximum of the Softmax probability (MSP) can be used as the monitoring
statistics, which represent how certain the neural network is about
the specific sample $\mbx$ is $\max_{k}f_{k}(\mbx)$. In other words,
a sample is detected to be out-of-distribution if its negative MSP
score is larger than a certain threshold $c$ as $-\max_{k}f_{k}(\mbx)>c$.

\subsubsection{Out-of-DIstribution detector for Neural networks (ODIN)}

The ODIN method was introduced in \citep{LiangLS18} as an improvement
over the MSP method. ODIN is composed of two additional improvements
to further improve its OOD Detection accuracy. The first improvement
is the temperature scaling, which was originally proposed by \citet{hintonDistilling2015}
for knowledge distillation of large neural networks and later used
to calibrate the classification uncertainty of the modern neural network
in \citep{guo2017calibration} to solve the overconfident issues in
large deep neural networks. The intuition is that by calibrating the
classification uncertainties, the MSP score could be more accurate.
The formulation of temperature scaling is defined as follows: for
a given input, temperature scaled output of the Softmax for a class
$k$ is formulated as below, with the single hyper-parameter being
the temperature $T$. The adjusted softmax function for class $k$
can be formulated as follows: 
\begin{equation}
f_{k}(\mbx;T)=\frac{\exp{(g_{k}(\mbx)/T})}{\sum_{j=1}^{K}\exp{(g_{j}(\mbx)/T)}}.\label{eq: temp-scaling}
\end{equation}

The second improvement to the original MSP method is the adversarial
perturbation. Recent research shows that most deep neural networks are not robust to adversarial perturbation \citep{GoodfellowSS14} and highlights the failure of some anomaly detection algorithms caused by adversarial attact\citep{biehler2024sage}.
ODIN method takes advantage of this property and claims that adversarial
perturbation trained for in-distribution samples will be far more
effective for the in-distribution samples compared to the out-of-distribution
samples, which further increases the MSP score difference for these
two groups. The formulation of adversarial perturbation is given as
follows: we first add adversarial perturbing into $\mbx$, and the
perturbed version of input is denoted by $\tilde{\mbx}$, which is
computed as follows: 
\begin{equation}
\Tilde{\mbx}=\mbx-\epsilon\cdot\text{sign}(-\nabla_{\mbx}\log f_{\hat{y}}(\mbx;T)),\label{eq: adv-purtabation}
\end{equation}
where $\hat{y}=\argmax_{k}f_{k}(\mbx;T)$ and $\epsilon$ is the perturbation
magnitude.

Similar to the MSP method, an OOD sample is detected if the negative
MSP score is larger than a threshold $c$, as shown below 
\begin{equation}
-\max f_{k}(\Tilde{\mbx};T)>c.
\end{equation}
Throughout the experiments in this study, $T$ and $\epsilon$ are
fixed to $1000$ and $0.0012$, respectively, as these are the values
are suggested by \citep{LiangLS18}.

\subsubsection{Deep Mahalanobis Detection}

Deep Mahalanobis Detection (DMD) was proposed in \citep{LeeLLS18} based on the scaled distance metric\citep{kim2011nonparametric}
and it has a different formulation than the aforementioned baseline
methods, given it is not based on the Softmax probability. It infers
a class-conditional Gaussian distribution over the outputs of the
penultimate layer $g(\mbx)$. For each class $k$, its respective
mean $\hat{\mu}_{k}$ and the global covariance $\hat{\Sigma}$ for
each class is calculated as below: 
\begin{align*}
\hat{\mu}_{k} & =\frac{1}{N_{k}}\sum_{\{n:y_{n}=k\}}g(\mbx_{n})\\
\hat{\Sigma} & =\frac{1}{N}\sum_{k}\sum_{\{n:y_{n}=k\}}(g(\mbx_{n})-\hat{\mu}_{k})(g(\mbx_{n})-\hat{\mu}_{k})^{T}
\end{align*}
where $N_{K}$ is the number of samples predicted as class $k$-based
on the MSP function, $y_{n}$ is the predicted label for sample $\mbx_{n}$.

The squared sample's Mahalanobis distance to class $k$ can be formulated
as:

\begin{equation}
D_{k}(\mbx)=(g(\mbx_{n})-\hat{\mu}_{k})^{T}\hat{\Sigma}^{-1}(g(\mbx_{n})-\hat{\mu}_{k}).
\end{equation}

Finally, for a given output and threshold $c$, the Minimum Mahalanobis
distance of the feature embedding $g(\mbx^{n})$ and the closest class
center $\hat{\mu}_{k}$ can be used as the OOD statistics and an OOD
sample is detected if and only if 
\begin{equation}
\min_{k}D_{k}(\mbx)>c.
\end{equation}

\subsection{Hierarchical Regularization in Hierarchical Classification \label{subsec:hierarchical-reg-anom-score}}

Typically, the classification objective is to minimize the total cross-entropy
loss over the training dataset by optimizing the parameters of the
neural network. To construct the cross-entropy loss, we first have
to define the one-hot embedding function. Assume a label $y_{n}$
from the dataset. The one-hot embedding $l_{k}:k\in\{1\dots K\}\to\{0,1\}^{K}$
maps a label to a $K$-length binary vector, where it attains only
one '1' on the dimension of the label $k$ and $K$ is the total number
of classes. In other words, $l_{k}(y^{n})=1$ if and only if $k=y_{n}$
and is zero otherwise.

The cross-entropy objective function over the training dataset $\mathcal{D}$
can be formalized as follows: 
\begin{equation}
\min_{\mbtheta}E(\mbtheta)=\min_{\mbtheta}-\sum_{n=1}^{N}\sum_{k=1}^{K}l_{k}(y_{n})\log f_{k}(\mbx_{n},\mbtheta)
\end{equation}

\begin{figure}[!t]
\centering \includegraphics[width=0.55\linewidth]{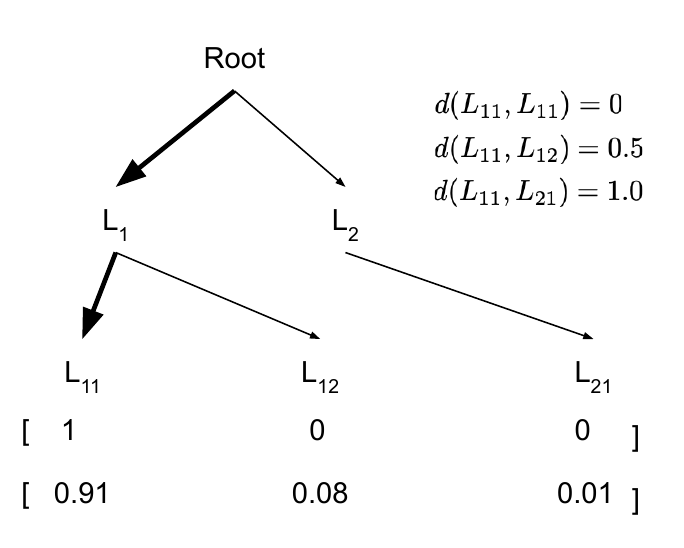}
\caption{Graphical illustration of the soft labeling logic. A hypothetical
two-level fault taxonomy is given. The distances show how the least common
ancestor-based distances manifest themselves for this structure. Using
these distances and \Cref{eq:soft-label}, and taking $\beta=5$,
we obtain the soft labels in the second row under the leaf labels,
given the real label is $L_{11}$. For comparison, one-hot labeling
is also shown for the same case.}
\label{fig:graphical-illustration-soft-labels} 
\end{figure}

Note that in this formulation, the model is only penalized for the
prediction made for the actual label. The problem is that when it
makes a mistake in predicting certain classes, it is completely indifferent
to the prediction of similar classes. This is the core challenge in
the hierarchical classification literature. In this paper, we will
use one of the state-of-the-art techniques from that literature for
hierarchical classification, namely the soft label formulation from
\citep{BertinettoBetterMistakes20} to model the relationship of labels.
The soft label is a label representation trick to transform the loss
function in a way that is sensitive to the predictions of the other
classes, too. Specifically, the new loss formulation forces the prediction
mistakes of the model to be hierarchically consistent. A closer look
at how the soft label embeddings are constructed is illustrated as
follows:

\begin{equation}
l_{k}^{\text{soft}}(i)=\frac{\exp(-\beta d(k,i))}{\sum_{j=1}^{K}\exp(-\beta d(j,i))}.\label{eq:soft-label}
\end{equation}
Here $\beta$ is a constant parameter, and $d(i,j)$ can be any proper
distance function that determines how close two labels are in the
hierarchy. Here, we choose 
to use the normalized lowest common ancestor distance, which is a
common measure of the distance between two nodes $i$ and $j$, on
a taxonomy tree $\mathcal{T}$ \citep{BertinettoBetterMistakes20}.
\begin{equation}
d(i,j)=\frac{LCA(i,j)}{h_{\mathcal{T}}}.\label{eq:distance-measure}
\end{equation}
Here, $LCA(i,j)$ is the lowest common ancestor, which is the lowest
ancestor that has both $i$ and $j$ as descendants. Furthermore,
to normalize this distance, we can divide the distance by the height
of the tree $h_{\mathcal{T}}$, so that the distance is normalized
to $1$. Note the label probabilities still sum up to one, but the
value assigned to the real label is amortized by a fraction, which
is redistributed to other classes in a hierarchically consistent manner.
See \Cref{fig:graphical-illustration-soft-labels} for an illustration
of the soft labeling mechanism. For example, in \Cref{fig:graphical-illustration-soft-labels},
the depth of the tree $h_{\mathcal{T}}=2$ and $LCA(L_{11},L_{12})=1$.
Therefore, $d(L_{11},L_{12})=\frac{LCA(L_{11},L_{12})}{h_{\mathcal{T}}}=0.5$.
Similarly, $LCA(L_{11},L_{21})=2$, therefore, $d(L_{11},L_{21})=\frac{LCA(L_{11},L_{21})}{h_{\mathcal{T}}}=1$.

Replacing one-hot embedding with soft labels, we obtain the updated
loss function as follows:

\begin{equation}
E(\mbtheta)=-\sum_{n=1}^{N}\sum_{k=1}^{K}l_{k}^{\text{soft}}(y_{n})\log f_{k}(\mbx_{n},\mbtheta)\label{eq:hier-reg-loss}
\end{equation}

Note that with the updated version of the loss function, the prediction
function $f(\mbx,\mbtheta)$ is not only encouraged to make the correct
prediction but also to do so in a way that reflects the hierarchical
relationships between the classes. We call this mechanism hierarchical
regularization. The strength of this regularization is determined
by the parameter $\beta$ in \Cref{eq:soft-label} where lower values
of $\beta$ assign greater importance to consistency.

\subsection{Proposed Methods \label{subsec:metho}}

In this section, we discuss the proposed methods in detail. We will
first demonstrate the hierarchically consistent score function in
\Cref{subsubsec: hierscore} and demonstrate how it can be
combined with three state-of-the-art OOD methods: MSP, ODIN, and DMD.
Finally, we would like to discuss the intuition behind the proposed
hierarchically consistent score.

\subsubsection{Proposed Hierarchically Consistent Score Function \label{subsubsec: hierscore}}

In this section, we will show how we can extend three established
methods in the literature using hierarchical regularization and also
propose a novel hierarchically consistent score function for anomaly
detection. The three methods are Maximum Softmax Probability (MSP)
\citep{HendrycksG17}, Out-of-DIstribution detector for neural networks
(ODIN) \citep{LiangLS18} and Deep Mahalanobis Detector (DMD) from
\citep{LeeLLS18}, introduced in \Cref{subsec: OODbasline}.

MSP and ODIN use the negative maximum softmax probability as an out-of-distribution
(OOD) score, and detect a sample $\mbx$ is out-of-distribution if
and only if 
\begin{equation}
-\max\{f_{k}(\mbx,\mbtheta^{*})|k\in\{1\dots K\}\}>c,\label{eq:max-softmax}
\end{equation}
where $\mbtheta^{*}$ are the parameters optimized over the training
set, and thus, they are fixed during test time, and $c$ is the threshold
for OOD detection.

Under the flat classification assumption, the formulation in \Cref{eq:max-softmax}
is reasonable, given that the model assumes that other classes are
equally unimportant given the predicted class, and there is no apparent
need to incorporate information outside the primarily predicted class.
However, we argue that if hierarchical regularization is used for
training, the OOD score function should be updated accordingly to
reflect hierarchical consistency. The most straightforward formulation
to obtain a weighted score would be using the soft labels. Let $\hat{y}=\argmax_{k\in\{1\dots K\}}\{f_{k}(\mbx,\mbtheta^{*})\}$
be the predicted label with the highest prediction score. Here, we
detect a sample $\mbx$ as an unknown fault if and only if 
\begin{equation}
-\sum_{k=1}^{K}l_{k}^{\text{soft}}(\hat{y})\log f_{k}(\mbx,\mbtheta^{*})>c,\label{eq:hier-const-score}
\end{equation}
where $l_{k}^{\text{soft}}(\hat{y})$ is the soft label defined in
\Cref{eq:soft-label} and $c$ is the threshold for the OOD detection.

The effectiveness of this formulation depends upon a fundamental assumption,
which is the samples generated from the in-distribution process yield
hierarchically consistent predictions because they are trained with
hierarchical regularization, whereas no such guarantee exists for
out-of-distribution samples. Therefore, the proposed OOD score is
able to detect OOD samples not only based on the maximum Softmax probability
but also on whether the prediction Softmax probability $f_{k}(\mbx,\mbtheta^{*})$
is hierarchically consistent. In our experiments, we will replace
\Cref{eq:max-softmax} with \Cref{eq:hier-const-score}, whenever
hierarchical regularization is used during training. Furthermore,
it is easy to show that when the predicted label distribution follows
the hierarchical structured defined by the soft labels, $f_{k}(\mbx,\mbtheta^{*})=l_{k}^{\text{soft}}(\hat{y})$,
the OOD score -$\sum_{k=1}^{K}l_{k}^{\text{soft}}(\hat{y})\log f_{k}(\mbx,\mbtheta^{*})$
can be minimized. Therefore, this detector aims to detect the outlier
samples that are hierarchically inconsistent.

\subsubsection{Insights to improve upon the state-of-the-art OOD Methods \label{subsec:Insights-to-improve}}
In this section, we delve into the insights explaining why hierarchical treatment can enhance the performance of state-of-the-art Out-of-Distribution (OOD) methods, including MSP, ODIN, and DMD. A summary of extensions to these base methods, both in the training and monitoring stages, is provided in \Cref{tab:summary}. We particularly focus on the improvements realized at the monitoring stage.

\noindent \textbf{Improvement over the flat MSP Method}

Maximum Softmax Probability (MSP) employs the maximum of the Softmax Probability, denoted as $\max_{k}f_{k}(x)$ for anomaly detection. In contrast, the hierarchical version of MSP utilizes \Cref{eq:hier-const-score}
for anomaly detection. As previously mentioned, the hierarchical consistent score aims to leverage the hierarchical structure to enhance the detection score. We elucidate the intuition behind this improvement through \Cref{Proposition-1: hiearchical-score}.

\begin{prop}
\label{Proposition-1: hiearchical-score} If we define $s(f_{k})=-\sum_{k=1}^{K}l_{k}^{\text{soft}}(\hat{y})\log f_{k}(\mbx,\mbtheta^{*})$
as a function of $f_{k}$, where $f_{k}$ is the output of the Softmax
layer, which satisfies $\sum_{k}f_{k}=1$. The hierarchically consistent
score $s(f_{k})$ will be minimized if and only if $f_{k}=l_{k}^{\text{soft}}(\hat{y})$. 
\end{prop}

The proof is straightforward and can be seen in \Cref{subsec:Proof-of-Proposition 1}.
 \Cref{Proposition-1: hiearchical-score} demonstrates that the proposed hierarchically consistent score will be minimized if and only if the Softmax output $f_{k}$ aligns with the soft labels. This is inherently hierarchically consistent, as the soft labels are defined based on the hierarchical tree structure $l_{k}^{\text{soft}}(\hat{y})$.
Therefore, test samples that violate hierarchical consistency will
yield in higher scores and will be flagged as unknown faults. 
The validation of this core assumption for the improvement of the MSP will be discussed in \Cref{subsec: improveMSP}. 

\noindent \textbf{Improvement over the flat ODIN Method}

The ODIN method enhances the original MSP in two significant ways: temperature scaling and adversarial perturbation.

\emph{\uline{Temperature scaling framework}} corrects the overconfident
Softmax score by incorporating a temperature term in \Cref{eq: temp-scaling}.
Using the soft label approach, the predicted probability aims to align with the soft label probability $l_{k}^{soft}(\hat{y})$.
\begin{prop}
\label{prop2: temp-scaling}When the temperature $T$ is large, $f_{k}(\boldsymbol{x};T)=\frac{1}{K-\frac{1}{T}\Sigma_{j\neq k}^K (g_k(\boldsymbol{x})-g_j(\boldsymbol{x}))}$
\end{prop}

The proof of \Cref{prop2: temp-scaling} is elaborated in \Cref{subsec:Proof-of-Proposition 2 }.
For in-distribution data, the $f_{k}(\boldsymbol{x};T)$ is more hierarchically
consistent, where $g_{\hat{y}}(\boldsymbol{x})$ is largest and other
$g_{k}(\boldsymbol{x})$ is smaller. For outliers, $g_{\hat{y}}(\boldsymbol{x})$
and $g_{k}(\boldsymbol{x})$ becomes much closer. 

\emph{\uline{Adversarial Perturbation}} is added to data $\boldsymbol{x}$
as shown in \Cref{eq: adv-purtabation}, and the perturbed version
of input is denoted by $\tilde{\boldsymbol{x}}$. The label is determined
by the original data $\boldsymbol{x}$ as $\hat{y}=\arg\max_{k}f_{k}(\boldsymbol{x};T)$.
However, the hierarchically consistent score is computed through the
perturbed data $s(f_{k})=-\sum_{k=1}^{K}l_{k}^{\text{soft}}(\hat{y})\log f_{k}(\tilde{\mbx})$.
We aim to provide intuition on why this perturbation can enhance the
OOD performance through the \Cref{prop3: perturbation}
and the subsequent remarks. 
\begin{prop}
\label{prop3: perturbation}The perturbation by $\tilde{\boldsymbol{x}}=\boldsymbol{x}-\epsilon\text{sign}(-\nabla_{\boldsymbol{x}}\log f_{\hat{y}}(\boldsymbol{x};T))$
leads to the following Taylor expansion of the hierarchically consistent
score $-\sum_{k}l_{k}^{\text{soft }}\left(\hat{y}\right)\log f_{k}\left(\tilde{\boldsymbol{x}}\right)=-\sum l_{k}^{\text{soft}}(\hat{y})\log f_{k}(\boldsymbol{x})+\epsilon U_{1}+\epsilon U_{2}+O(\epsilon^{2})$,
where $U_{1}=-l_{\hat{y}}^{\text{soft}}(\hat{y})\|\nabla_{\mbx}\log f_{\hat{y}}(x)\|_{1}$
and $U_{2}=-\sum_{k\neq\hat{y}}l_{k}^{\text{soft}}(\hat{y})\text{sign}(-\nabla_{\mbx}\log f_{\hat{y}}(x))\cdot\nabla_{\mbx}\log f_{k}(\boldsymbol{x})$ and
$U_{2}\geq -\sum_{k\neq\hat{y}}l_{k}^{\text{soft}}(k)\|\nabla_{\mbx}\log f_{k}(x)\|_{1}$. 
\end{prop}

The proof of \Cref{prop3: perturbation} is shown in \Cref{subsec:Proof-of-Proposition 3}.
\Cref{prop3: perturbation} demonstrated that the first-order effect of the perturbation on the proposed hierarchically consistent
score can be decomposed into two parts $U_{1}$ and $U_{2}$. $U_{1}$
illustrates the effect of the label $\hat{y}$ itself, while $U_{2}$
represents the effects of other labels $k$ on the label $\hat{y}$.
In general, $U_{2}=0$ if there is no hierarchical structure on the
labels. From \Cref{prop3: perturbation}, we would like
to illustrate the following two important observations. 
\begin{rem}

\label{rem: U1} $U_{1}$ is typically  smaller for in-distribution
data compared to outliers. 
\end{rem}

\Cref{rem: U1} is demonstrated in the original paper \citep{LiangLS18}. For more validation of how $U_1$ and $U_2$ differ the abnormal samples from the normal samples, please refer to \Cref{subsec:U1U2example} for the validation using the real case study in \Cref{sec: case study}.
 This behavior leads to
a better detection performance of ODIN compared to the baseline MSP
methods. 
\begin{rem}
\label{rem: U2} $U_{2}$ is typically smaller for in-distribution
data with hierarchical structures compared to outliers. 
\end{rem}

\Cref{subsec:U1U2example} provides an illustrative example for \Cref{rem: U2} using our dataset. We will briefly illustrate the insight of \Cref{rem: U2} here.
The reason is that for in-distribution data with hierarchical structures,
there exists a set of classes $k\in\mathcal{K}$ that is close to $\hat{y}$
where $l_{k}^{\text{soft}}(k)$ is large. For these $k\in\mathcal{K}$,
$\text{sign}(-\nabla_{x}\log f_{\hat{y}}(x))\approx\text{sign}(-\nabla_{x}\log f_{k}(x))$,
given these two classes are similar (or hierarchically consistent). Therefore,
for in-distribution data, $U_{2}$ can reach the lower bound $\sum_{k\neq\hat{y}}l_{k}^{\text{soft}}(k)\|\nabla_{x}\log f_{k}(x)\|_{1}$
much easier. On the other hand, for outliers, $\text{sign}(-\nabla_{x}\log f_{\hat{y}}(x))$
will be random and can have different signs compared to $\nabla_{x}\log f_{k}(x)$.
Therefore the inner product $\text{sign}(-\nabla_{x}\log f_{\hat{y}}(x))\cdot\nabla_{x}\log f_{k}(x)$
will be much smaller or even becomes $0$ in $U_{2}$. 

In conclusion, \Cref{rem: U1} and \Cref{rem: U2} guarantee
that the adversarial perturbation will increase the hierarchical consistent
score much larger for in-distribution data compared to outliers. This core assumption of the improvement over the ODIN methods will be validated in \Cref{subsec: improveODIN}.

\noindent \textbf{Improvement over the flat DMD method}

The DMD method does not necessarily require additional treatment
for its OOD score formulation. By definition, it fits a class-conditional
Gaussian distribution over the entire prediction output of a neural
network, not just the maximum prediction score. However, hierarchical consistency
will tighten the spread of the conditional distributions of each class,
especially if these two classes belong to the same parent.  This should result in more conservative anomaly thresholds, thereby reducing Type-I errors. 
In our experiments, we will employ a simpler variant of this method, fitting distributions only on the outputs of the penultimate layer. For further details, readers are referred to Equation 1 and Equation 2
in \citep{LeeLLS18}.

The numerical and real-case study validation of these core assumptions for DMD improvement will be discussed in \Cref{subsec: improveDMD}.

\begin{table}
\caption{Summary of proposed extensions over base methods, at training and
at the monitoring stage.}
\centering %
\begin{tabular}{lll}
\toprule 
Base Method  & At Training  & At Monitoring \tabularnewline
\midrule 
MSP  & \Cref{eq:hier-reg-loss}  & \Cref{eq:hier-const-score} \tabularnewline
ODIN  & \Cref{eq:hier-reg-loss}  & \Cref{eq:hier-const-score} \tabularnewline
DMD  & \Cref{eq:hier-reg-loss}  & --- \tabularnewline
\bottomrule
\end{tabular}\label{tab:summary} 
\end{table}

\subsection{Practical Guidelines and Tuning Parameter Selections \label{subsec:tuning}}

\subsubsection{Parameter $\beta$ Selection}

The parameter $\beta$ determines the importance of the hierarchical
information in training and testing, and therefore, we want to explore
its impact on our methodology. When the value of $\beta$ is small,
the methodology places less emphasis on the knowledge of hierarchical
structures. As $\beta\rightarrow0$, the soft-label embedding in \Cref{eq:soft-label}
will be a uniform distribution, which makes the label useless. If
$\beta$ is large, the proposed hierarchically consistent OOD score
will ignore any prior assumptions or knowledge about the hierarchical
information and becomes the standard Maximum Softmax Probability as
illustrated in Proposition 2. 
\begin{prop}
\label{prop: msp}When $\beta\rightarrow+\infty$, for $\mbx$, the
hierarchical consistent score can be used to detect anomalies as: 
\[
-\sum_{k=1}^{K}l_{k}^{\text{soft}}(\hat{y})\log f_{k}(\mbx,\mbtheta^{*})>c
\]
is equivalent to $\max_{k}\{f_{k}(\mbx,\mbtheta^{*})\}<c'$, where
$c'$ is a positive constant threshold. This implies that the proposed
hierarchically consistent OOD score will become the traditional Maximum
Softmax Probability (MSP) criterion used in MSP and ODIN methods when
$\beta\rightarrow+\infty.$
\end{prop}

The proof is given in \Cref{subsec:Proof-of-Proposition 6}.
 \Cref{prop: msp} shows the connection between the proposed
hierarchically consistent score and the MSP. In summary, it reveals
that the proposed hierarchically consistency score will become equivalent
to the detection score in the traditional MSP method when $\beta\rightarrow+\infty$.
In practice, we may need to select a proper value of $\beta$ to incorporate
the hierarchical structure information. 

\subsubsection{OOD Detection Procedure and Threshold c}

\begin{figure}[!t]
\centering \includegraphics[width=0.65\linewidth]{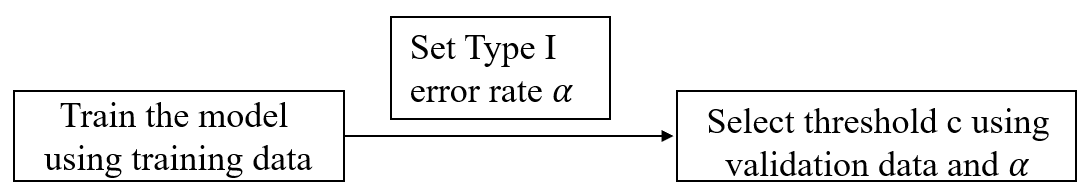}
\caption{Example flowchart for critical value}
\label{fig:flow_threshold} 
\end{figure}

In the context of industrial application, our proposed method can
be effectively applied by the following steps: 
\begin{enumerate}
\item In data preparation, obtain in control dataset $\mathcal{D}$ and
partition it to two sets: training and validation sets $\mathcal{D}_{trn}$
and $\mathcal{D}_{val}$. 
\item Train the hierarchical classifiers using the training data alone $\mathcal{D}_{trn}$. 
\item Computing the testing statistic for all validation samples and takes
its $1-\alpha$ percentile as $c$. Remove the out-of-control process
and update the percentile until convergence. 
\item Compute the test statistics for new samples and start to identify
samples as out-of-control if they are above the threshold $c$ 
\end{enumerate}
In practice, some evaluation metrics, such as the type I error
of the classification results in the validation set can be used to
obtain a proper threshold $c$ depending on different industrial applications.
For example, we can try to use validation data to get the critical
values for different methods via the process shown in \Cref{fig:flow_threshold}.

\section{Case Study \label{sec: case study}}

In this section, we implement the proposed methodology in real word
data as described in \Cref{subsec:rolling-dataset} and introduce
the details of the experimental design and the evaluation criteria
in \Cref{subsec:experiment-design and eva}. Finally, the result
comparison is presented in \Cref{sec:results}.

\subsection{Hot Steel Rolling Dataset\label{subsec:rolling-dataset}}

Our proposed extensions will be evaluated on a real-life case study
image dataset collected from a hot steel rolling process. There are,
in total, 3732 image patches that have been detected as potential
anomalies obtained by an industrial vision monitoring system. These
images have been carefully labeled by domain engineers. Each defect
is assigned a two-level label, where there are 8 major categories
and 13 subcategories. Please refer to \Cref{fig:graphical-illustration-hierarchy-rolling}
for the full tree of defects hierarchy for this dataset, as well as
example images for each leaf category.

The following details of the dataset are provided: 1) Image dimension:
The original image sizes for different images could have different
dimensions. To make them as the uniform input size $224\times224$,
we first center crop the figure into an $80\times80$ square and resize
them to the required size; 3) They are all grayscale images; 4) Sample
sizes: 18 to 135 samples are available for different defect types.
The detailed sample size table is listed in \Cref{app:samplesize}.

For this study, we randomly partition the data set into train, validation,
and test partitions with sizes $60\%-20\%-20\%$ of the original dataset,
respectively. The proportions of class prevalence remain the same
across all partitions. In other words, we employ a stratified split
based on class proportions.

\subsection{Experimental Design and Evaluation Criteria \label{subsec:experiment-design and eva}}

We design an experiment in which we control all possible aspects of
a neural network training process except for the hierarchical regularization
in \Cref{subsec:hierarchical-reg-anom-score} and unknown OOD score
functions discussed in \Cref{subsec:metho}. With this design, we
aim to answer the question of whether detection performance improves
considering hierarchical information of labels.

To simulate the emergence of a new class, we leave out a class from
the tree for each experiment. Singling out of each class can be considered
a new scenario. There will also be a set of testing samples from known
classes to evaluate how the scoring differs between known and unknown
classes.

An overview of the steps of each experiment can be listed as follows: 
\begin{enumerate}
\item Leave out one of the classes as the emerging anomaly class. Only collect
the anomaly sample in a single test partition. 
\item Train a deep convolutional neural network (CNN) for the classification
task with samples in the training set over a set of predetermined
hyperparameters. 
\item Choose the CNN model with the optimal hyperparameter that produces
the best loss in the validation set, freeze its parameters, and proceed
with it to the testing stage. 
\item Compute the anomaly score function for the validation and testing
samples. 
\item Report summary evaluation statistics over all test samples. 
\end{enumerate}
The experiments are factorized in steps (2) and (4). In step (2),
we will compare the training with or without hierarchical regularization.
In step (4), we use MSP, ODIN, or DMD as explained in \Cref{subsec:hierarchical-reg-anom-score}.
For MSP and ODIN, the hierarchically consistent OOD scoring will be
used in step (4).

In step (1), we replicate the same experiment for four different scenarios.
In all scenarios, we remove all instances from one of the classes
in the training samples from A12, A31, A61 and A40. From the hierarchical
structure tree as shown in \Cref{fig:graphical-illustration-hierarchy-rolling},
there is 1) only one parent node having three child fault types, 2)
four parent nodes having two child fault types, 3) three parent nodes
having only one child fault type. In the experiment setting, we aim
to cover all these three scenarios in picking up our OOD class for
testing.

\begin{figure}[!t]
\centering \includegraphics[width=0.7\linewidth]{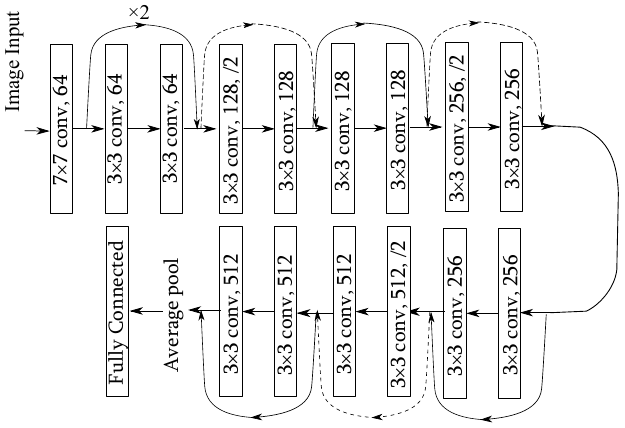}
\caption{ResNet18 Architecture \citep{HeZRS16}.}
\label{fig:resnet} 
\end{figure}

In the first scenario, we select class A12 as the novel fault to observe
the behavior of the detectors. This selection can represent the model's
performance on a class with two sibling leave nodes, and we assume
that hierarchical algorithms can help improve the detection of this
novel fault class (see \Cref{fig:graphical-illustration-hierarchy-rolling}).
For the next two scenarios, we select classes A31 and A61 as the novel
fault classes. Both classes have only one sibling leave node. In the
last scenario, we select class A40 as the novel fault class, as it
has no sibling leave node in the structure. Compared to other non-sibling
fault classes, the dissimilarity of this fault from all others is
relatively distinguishable. To evaluate the efficacy of our proposed
method, we have deliberately chosen this diverse range of novel fault
classes. Our intention is to ensure that our approach is robust and
capable of detecting novel faults in various scenarios. In all of
the experiment runs, we utilize the 18-layer variant of the ResNet
architecture for the classifier \citep{HeZRS16}. The detailed architecture
is given in \Cref{fig:resnet}.

As advised by \citet{HendrycksG17} and employed by most other papers
in the literature, we use Area Under the Receiving Operating Characteristics
curve (AUROC) given that it doesn't need to involve the careful selection
of the threshold $c$ and type-I error. Here, in literature, the AUROC
can be interpreted as the likelihood of a detector to score a random
sample from an unknown fault higher than that is randomly picked from
known fault types.

\subsection{Results \& Discussion \label{sec:results}}

In this section, we will discuss the comparison of the detection performance
of the hierarchical method with the flat method for all three aforementioned
benchmark methods, including MSP, ODIN, and DMD in all four different
scenarios.

\subsubsection{Detection Performance Comparisons}

\begin{figure*}[!t]
\centering \includegraphics[width=0.995\linewidth]{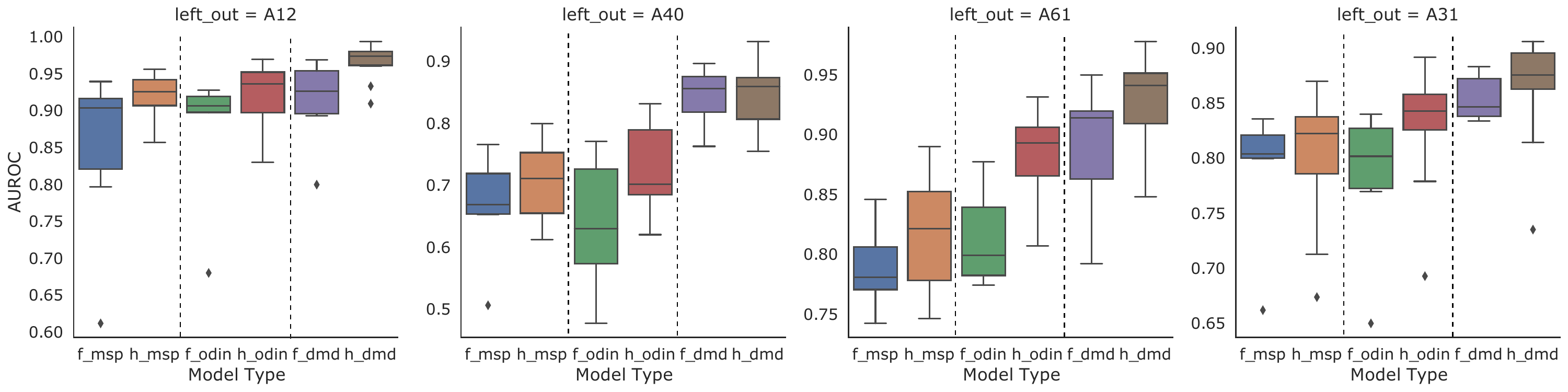}
\caption{AUROC Comparison: This figure includes all experiments with $\beta=10$
or $\beta=100$. Each subplot shows six values on the x-axis, representing
the model type and the baseline method used. Specifically, the prefix
\texttt{f\_} denotes \texttt{flat model} while \texttt{h\_} denotes
\texttt{hierarchical model}. The baseline methods \texttt{msp}, \texttt{odin},
and \texttt{dmd}, which correspond to the MSP, ODIN, and DMD methods
reviewed in \Cref{sec:methodology}, are also represented on
the x-axis. To aid in visual clarity, dotted lines are used to separate
the results of different baseline methods for each scenario in each
subplot. }
\label{fig:all_compare_high} 
\end{figure*}

In order to compare the performance via Area Under ROC Curve (AUROC)
between flat model and hierarchical model clearly, box-plots are shown
in \Cref{fig:all_compare_high} given proper $\beta$ values.
Further sensitivity analysis regarding the hyper-parameter $\beta$
will be investigated in the subsequent subsection. The evaluation
dataset of our models encompasses the complete test set, including
the known and unknown fault classes, across multiple replications
with different learning rates and train seed configurations. 

Regarding the model based on MSP, we can refer to the \texttt{f\_msp}
and \texttt{h\_msp} to find the performance comparison between flat
and hierarchical models. We first observe a relatively high novel
fault class detection AUC score for `A12' with AUROC larger than $0.9$,
given that this class is fairly different from the existing classes.
For example, `A12' has a black line within the defect region, and
`A11' and `A10' have white defective regions. `A31' and `A61' have
detection AUROC around 0.8, while `A40' has AUROC about 0.7. Compared
to other classes, `A40' has a very small sample size as shown in
\Cref{t:novel_sample_size} and may affect the test accuracy robustness.
We then observe an increase in both median and highest detection performance
in all four left-out scenarios when the hierarchical regularization
and consistency are employed at training and testing time, respectively.
\begin{table}[h]
\centering \caption{Sample size summary for left-out classes}
\begin{tabular}{|c|c|c|}
\hline 
Label    & Sample Size \tabularnewline
\hline 
A12   & 75 \tabularnewline
\hline 
A31    & 115 \tabularnewline
\hline 
A40    & 18 \tabularnewline
\hline 
A61  & 75 \tabularnewline
\hline 
\end{tabular}\label{t:novel_sample_size}
\end{table}

Regarding ODIN and hierarchical ODIN performance,we can refer to \texttt{f\_odin}
and \texttt{h\_odin} in \Cref{fig:all_compare_high} as well.
First, we have observed an improvement in the baseline detection performance
of ODIN over the MSP detector in most classes except class `A40' with
18 samples in testing, which has been demonstrated in \citep{LiangLS18}.
Second, we have observed a similar performance increase when the hierarchical
model structure is considered. This is not a surprising result, given
that both ODIN and MSP use the maximum softmax probability. More specifically,
the performance improvement of `A61' is the largest when ODIN is used
as the base detector compared to MSP, given that the temperature scaling
helps reduce the over-confident issue in the neural network and helps
achieve a more reasonable hierarchical consistent score.

Finally, we will show the novel fault detection result when DMD is
employed as the baseline detection. We also observe results strongly
in favor of employing a hierarchical treatment to increase the novel
fault detection performance. This reinforces the confidence in the
results as DMD employs a rather different strategy at detection time
as opposed to MSP and ODIN. As mentioned in \Cref{sec:methodology},
it is based on the feature representation without a softmax layer.
Based on the above observation and overall review of \Cref{fig:all_compare_high},
we can find the promising improvement of the hierarchical classifiers
in novel fault detection problems even without the re-design of the
deep learning architecture.

\subsubsection{Sensitivity Analysis}

\begin{figure*}[!t]
\centering \includegraphics[width=0.75\linewidth]{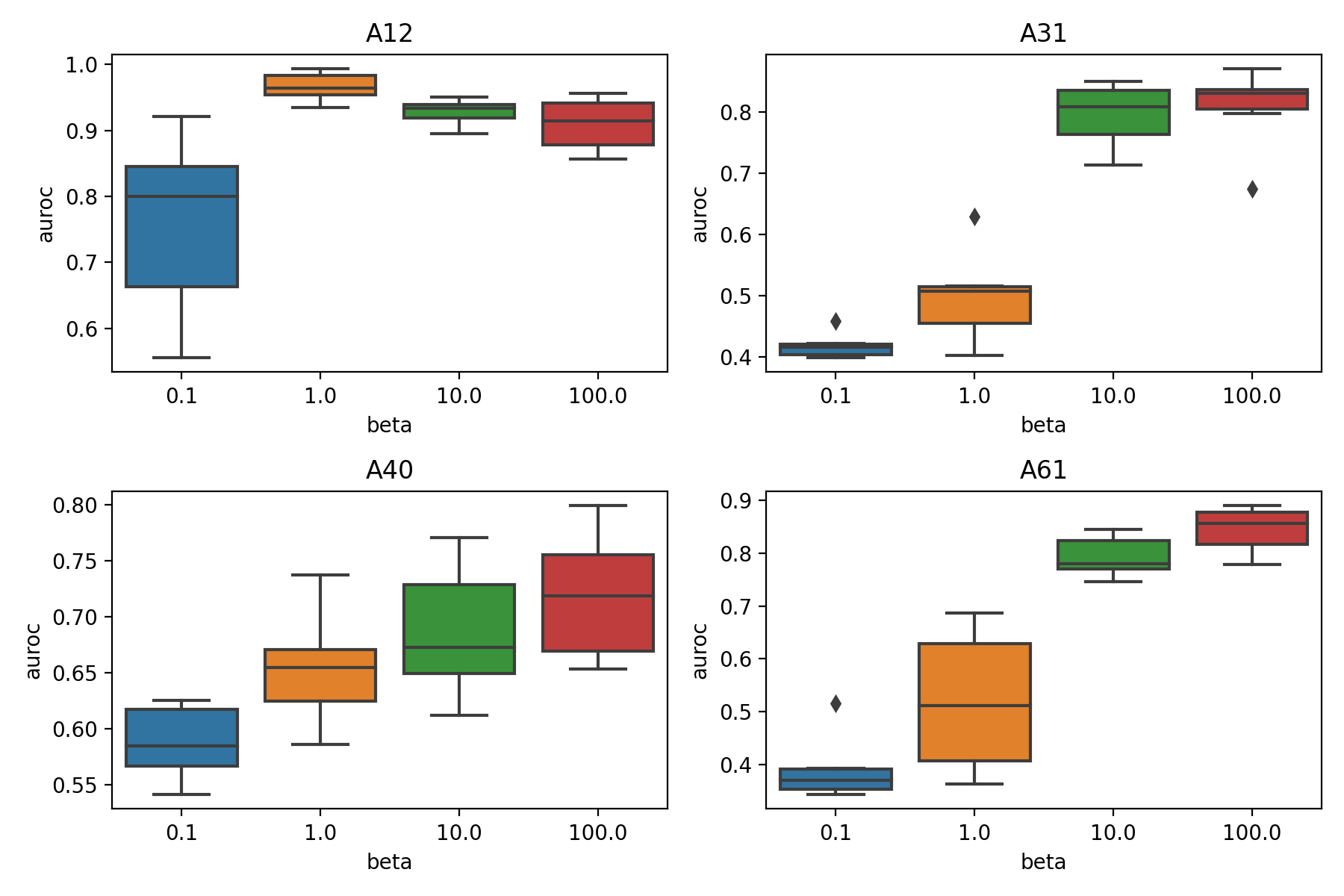}
\caption{Sensitivity analysis for hierarchical MSP model on different $\beta$}
\label{fig:msp_sensi} 
\end{figure*}

\begin{figure*}[!t]
\centering \includegraphics[width=0.75\linewidth]{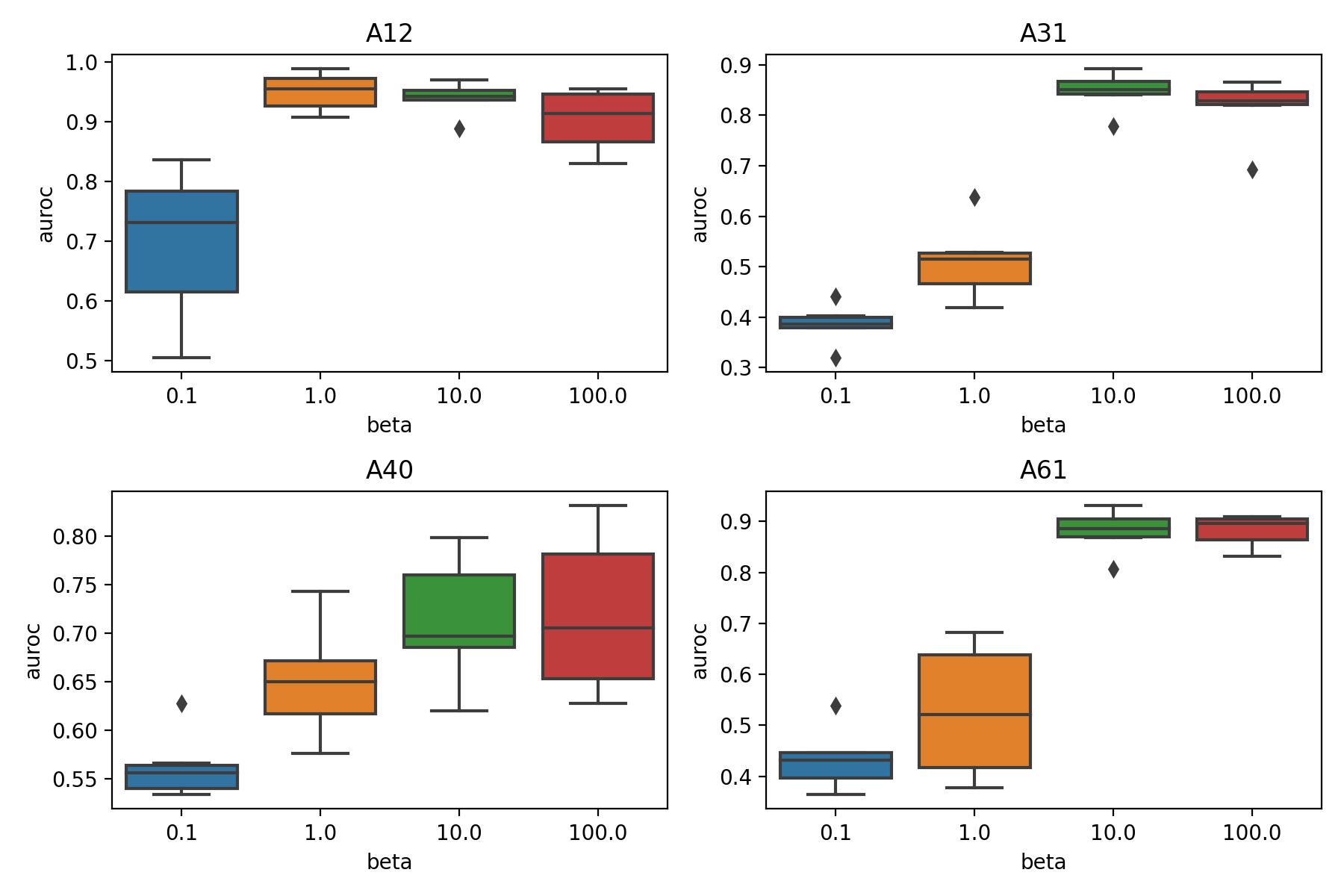}
\caption{Sensitivity analysis for hierarchical ODIN model on different $\beta$}
\label{fig:odin_sensi} 
\end{figure*}

\begin{figure*}[!t]
\centering \includegraphics[width=0.75\linewidth]{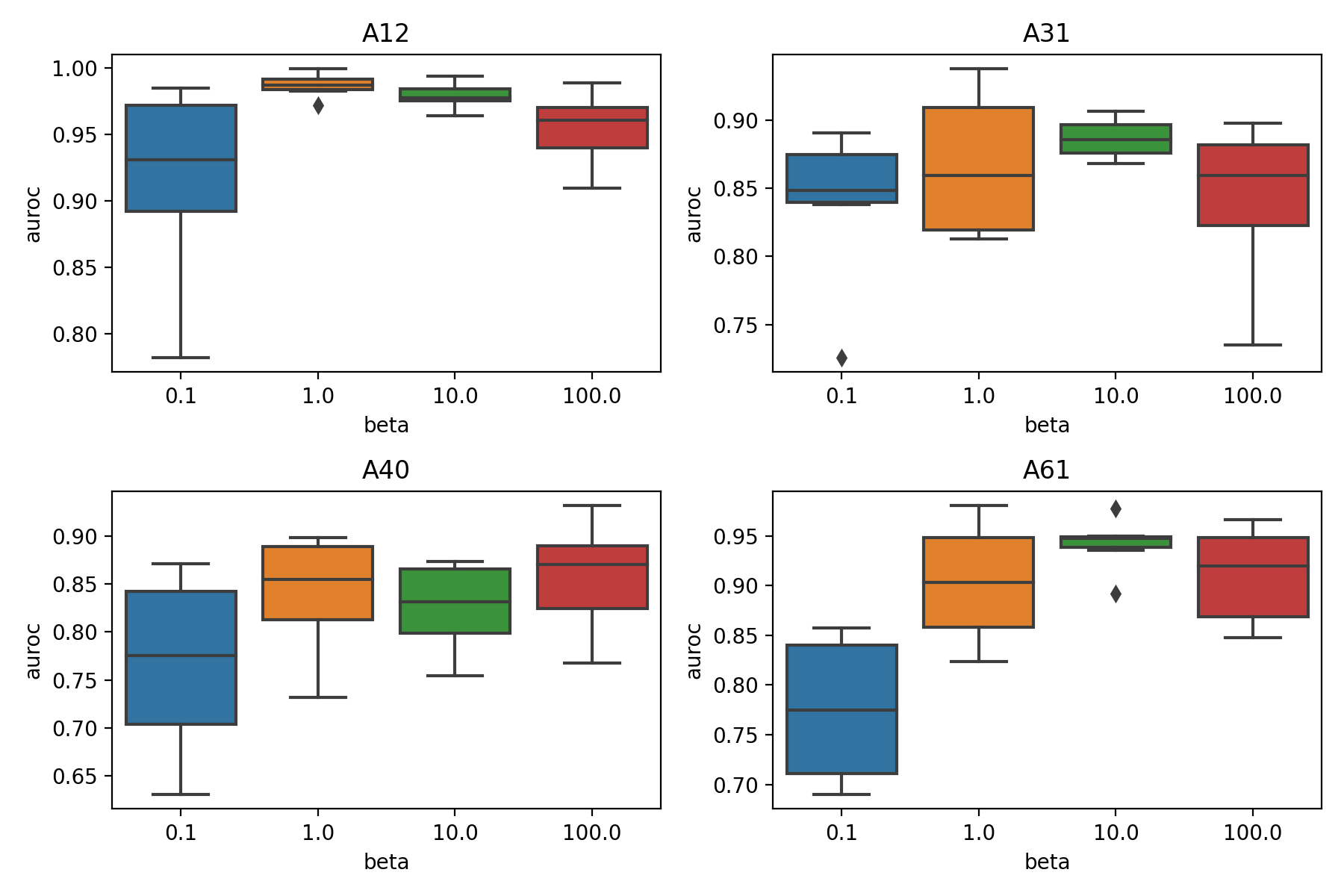}
\caption{Sensitivity analysis for hierarchical DMD model on different $\beta$}
\label{fig:dmd_sensi} 
\end{figure*}

To investigate the sensitivity of the model performance concerning
alternations in the soft-label embedding hyper-parameter $\beta$,
we have conducted a series of experiments in which we varied the values
of $\beta$ and recorded their impact on the model's performance.
For each experiment, $\beta$ is selected from the set $\beta\in\{0.1,1,10,100\}$.
\Cref{fig:msp_sensi}, \Cref{fig:odin_sensi}, and 
\Cref{fig:dmd_sensi} reflect the hierarchical model performance built
on different baseline methods.

First, through an examination of \Cref{fig:msp_sensi}, the
results generated from the model with $\beta=0.1$ are the poorest
among all scenarios, while models with $\beta\in\{10,100\}$ perform
fairly well in the MSP hierarchical model, except for the A12 scenario.
Thus, based on this observation, we can conclude that a very low value
of $\beta$ could negatively impact the hierarchical model performance.

\Cref{fig:odin_sensi} illustrates the sensitivity of ODIN hierarchical
models with respect to $\beta$. The outcomes of this sensitivity
analysis exhibit a closely similar pattern to that observed in the
previously discussed results of MSP hierarchical models. Specifically,
except for A12, models with $\beta\in\{10,100\}$ perform fairly well
in the ODIN hierarchical model. Additionally, models with $\beta=10$
perform very consistently over all classes. Thus, based on this observation,
we also can conclude that a very low value of $\beta$ could negatively
impact the hierarchical performance, and $\beta$ around 10 could
be a good option in parameter selection.

Upon examination of \Cref{fig:dmd_sensi}, we can also observe
the poor performance of the hierarchical model with $\beta=0.1$.
Conversely, hierarchical DMD models with $\beta\in\{1,10,100\}$ exhibit
levels of performance that are comparable to one another, demonstrating
low sensitivity across different values of $\beta$ between $[1,100]$.

After the examination of these visualizations of the sensitivity analysis
over hyper-parameter $\beta$, we can make a brief conclusion that
extremely low values of $\beta$ are sub-optimal for all the three
methods. Instead, selecting a value of $\beta$ from the set ${10,100}$
yields relatively stable and satisfactory model performance. Furthermore,
among these 4 different selections for $\beta$, our experimental
results suggest that $\beta=10$ is associated with the most stable
and optimal performance of the proposed hierarchical models. 

\subsection{Validation of Core Assumptions \label{subsec:Validation-of-Core}}

In this section, we would like to check the core assumptions that
hierarchically consistent scores can improve upon many state-of-the-art
OOD methods, including MSP, ODIN, and DMD. The insight into why hierarchically
consistent score works have been discussed in \Cref{subsec:Insights-to-improve},
which we would like to validate the major assumption that we made.

\subsubsection{Validation of Hierarchically consistent score improves upon MSP \label{subsec: improveMSP}}

\begin{figure*}[!t]
\centering \includegraphics[width=0.95\linewidth]{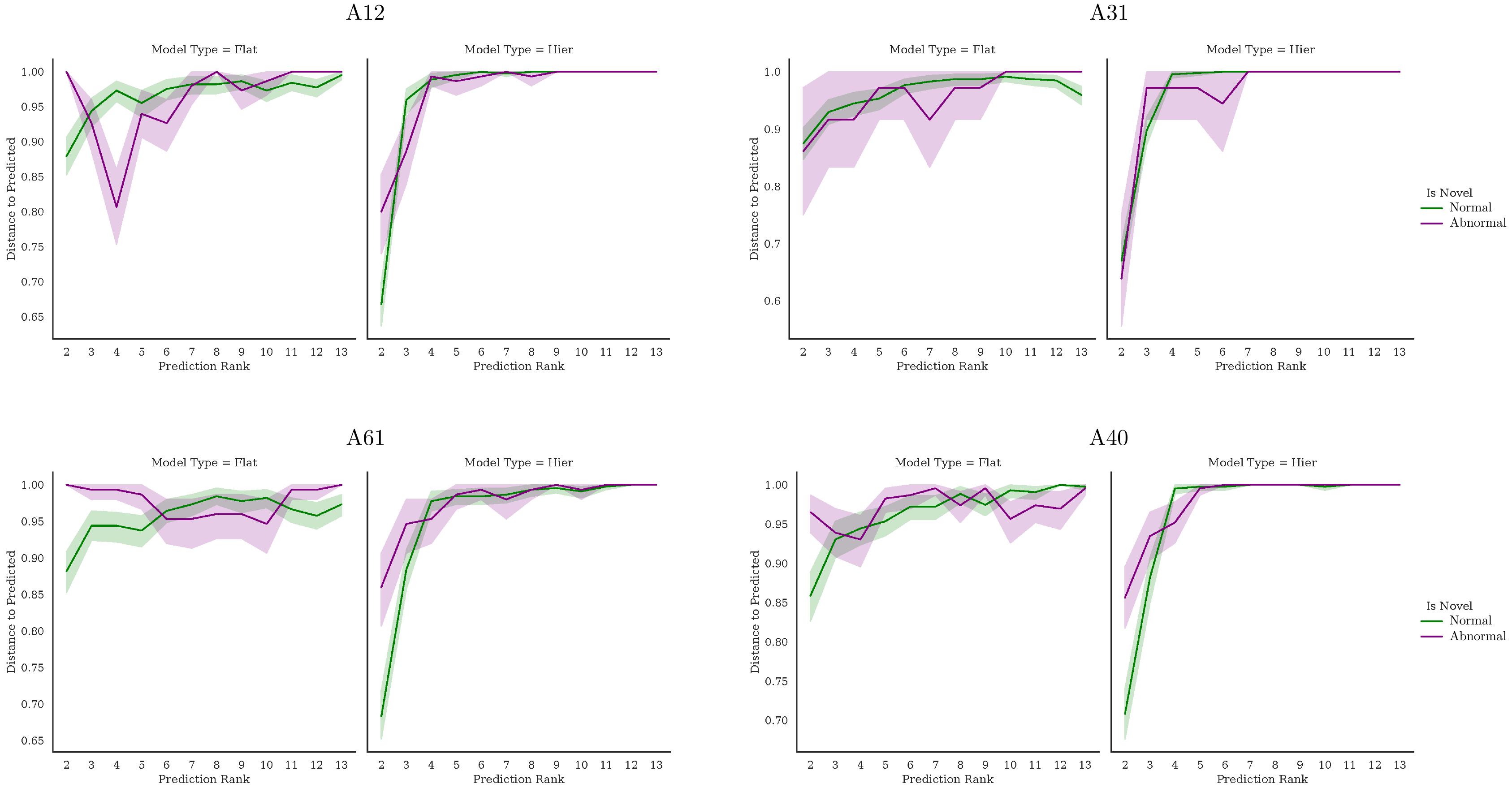}
\caption[Evolution of distance to predicted by prediction rank.]{Evolution of distance to predicted by prediction rank. Each subfigure
is created by an exemplar run from a scenario where one of the leaf
classes is left out. For each scenario, each element in a test is assigned
a prediction. The predictions are ranked and their lowest common
ancestor-based distance is recorded. The curves depict the evolution
of distances with increasing rank with a 95\% confidence interval band.
The line color denotes whether samples from known(normal) or unknown(abnormal)
faults were used to create summary statistics. Figures on the left
denoted when hierarchical regularization was not applied while for
the figures on the right, it was.}
\label{fig:rank_dist}
\end{figure*}

To validate the core assumption outlined in \Cref{subsec:hierarchical-reg-anom-score},
we present \Cref{fig:rank_dist}, which illustrates the relationship
between the ``distance to prediction'' and ``the prediction rank''.
The term ``distribution to prediction'' refers to the normalized lowest
common ancestor distance of the predicted label, as defined in \Cref{eq:distance-measure}.
The ``prediction rank'' represents the ranked Softmax probability.
We start counting from rank 2 on the x-axis, as the
``distribution to prediction'' is always $0$ for the predicted
label (i.e., the rank one probability of the Softmax probability); the
``distance to prediction'' signifies the distance from a given label
in the prediction to the true label. The experiments were
conducted over 300 iterations and the mean and 95\% confidence interval are presented for all four scenarios both for anomaly/novel fault samples (i.e., shown in purple) and normal/known samples (i.e., shown in green). 

The objective is to investigate whether a monotonic relationship exists between the ``distance to
prediction'' and ``the prediction rank''. If such a relationship is observed, it would imply that the mean distance to the predicted label
(i.e., Rank 1 prediction label) in the hierarchy increases monotonically
as the prediction rank decreases. This would further indicate that the prediction is ``hierarchically consistent''.

From this study, we can draw several interesting conclusions:
 1) In the model without the hierarchical treatment, the prediction
is not ``hierarchically consistent''. This suggests that a hierarchical
structure, if not explicitly enforced, is not inherently present in
traditional flat detection methods. For the model with the hierarchical
treatment, only the normal samples are hierarchically consistent.
However, for the anomaly samples, the monotonicity breaks on a few
ranks. 2) For the flat MSP, both the
normal and abnormal samples are not clearly separable. However,
when hierarchical treatment is applied
during model training, the separation between the curves becomes more distinct. This is particularly important for the proposed hierarchically
consistent anomaly score function, as it relies on this hierarchical
consistency to distinguish between normal and abnormal samples.
Specifically, the curve representing anomaly samples diverges from the curve for normal samples at earlier ranks rather than later ones.
Given that the hierarchically consistent score function places greater
emphasis on earlier ranks, this behavior further ensures the separability of anomaly samples. This is crucial for the success of MSP and ODIN, as both methods depend on this score function for effective anomaly detection.

\subsubsection{Validation of Hierarchically Consistent Score Improves upon ODIN \label{subsec: improveODIN}}

\begin{figure*}[!t]
\centering \includegraphics[width=0.8\linewidth]{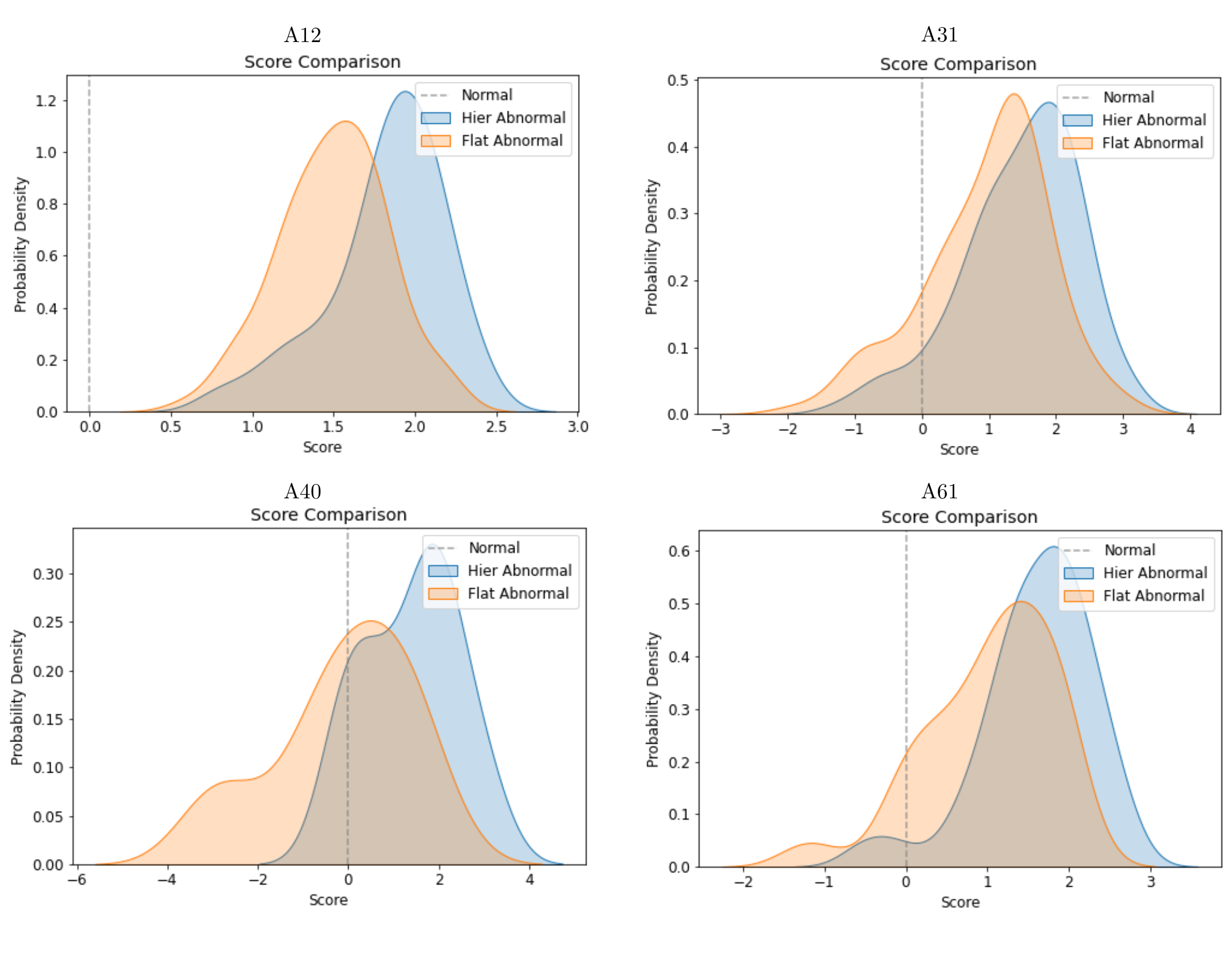}
\caption{Abnormal class score comparison for flat and hierarchical ODIN: 1) The grey dotted lines in subplots indicate that after standardization, the normal class score is near $0$. 2) Subplots with different colors indicate the standardized hierarchical consistency score comparison for 4 different left-out classes. Higher scores signify a greater likelihood of an instance belonging to the unknown/abnormal class.}
\label{fig:score_compare_ODIN_flat_hier}
\end{figure*}

The ODIN model, an enhanced version of the Maximum Softmax Probability (MSP) method, incorporates temperature scaling and perturbation techniques to address issues of overconfidence and improve the separation between normal and abnormal samples. In this study, we extend the ODIN model by incorporating hierarchical structure information and employing a hierarchical consistency score for both training and prediction. Specifically, we examine whether the hierarchical training approach can make ODIN more effective in distinguishing between normal and abnormal instances.  In \Cref{fig:score_compare_ODIN_flat_hier} we compare the performance of the newly proposed Hierarchically Trained ODIN model with that of the flat ODIN model, while maintaining consistent hyper-parameter settings. All the scores are standardized using the mean and standard deviation from the normal validation samples.

From \Cref{fig:score_compare_ODIN_flat_hier}, the abnormal class scores from hierarchically trained models (hier) validate the enhancement achieved by incorporating hierarchical structure into ODIN. 
The 'hier' model's abnormal class scores exhibit greater deviation from the standardized normal class scores, centered at zero, when compared to 'flat' model abnormal class scores. This enhancement underscores the utility of the hierarchical approach in bolstering the precision of out-of-distribution detection and showcases its potential to advance the state-of-the-art in anomaly detection models.

\subsubsection{Validation of Hierarchically consistent score improves upon DMD \label{subsec: improveDMD}}

\begin{figure*}[!t]
\centering \includegraphics[width=0.85\linewidth]{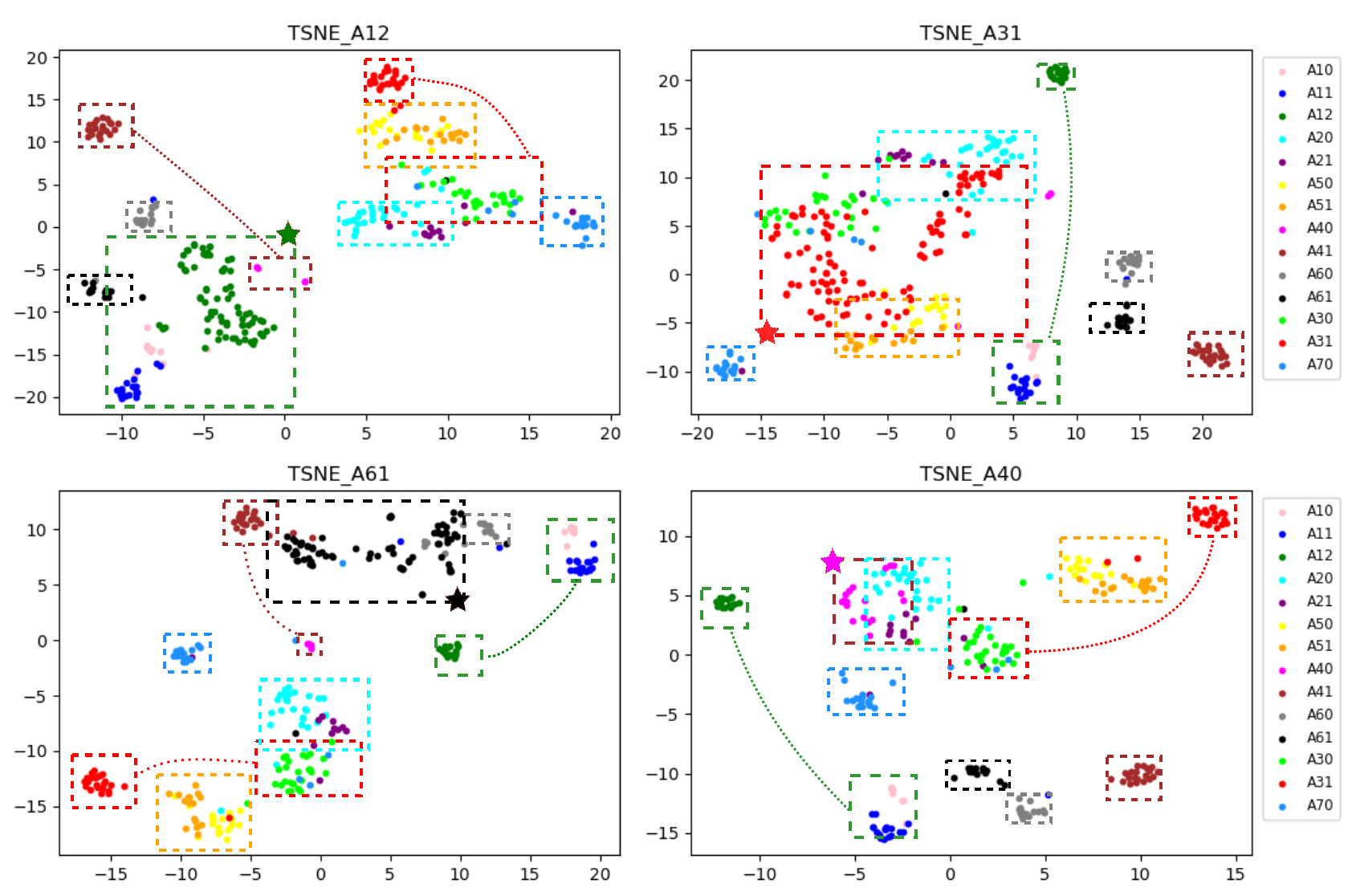}
\caption{t-SNE plots generated by baseline model: In each subplot, a dashed
box roughly squares the dots under the same first-level class (parent
class), and the star with the corresponding color denotes the left-out
class in that experiment.}
\label{fig:dmd_tsne_flat}
\end{figure*}

\begin{figure*}[!t]
\centering \includegraphics[width=0.85\linewidth]{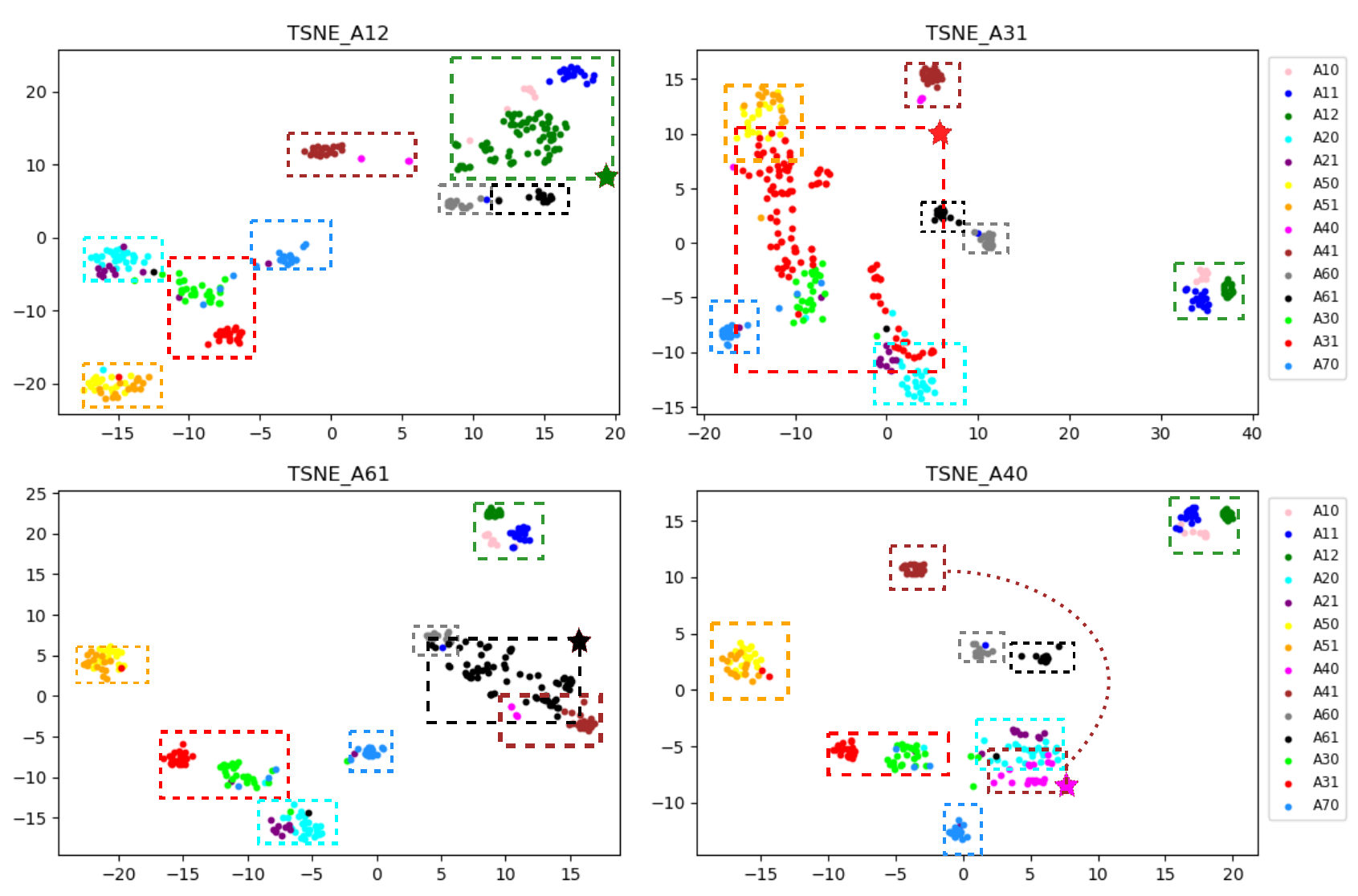}
\caption{t-SNE plots generated by proposed hierarchical model: In each subplot,
a dashed box roughly squares the dots under the same first-level class
(parent class). The star with the corresponding color denotes
the left-out class in that experiment.}
\label{fig:dmd_tsne_hier}
\end{figure*}

\Cref{fig:dmd_tsne_flat} and \Cref{fig:dmd_tsne_hier}
present the t-Distributed Stochastic Neighbor Embedding (t-SNE) plots
\citep{maaten_visualizing_2008} of the flat and hierarchical
classifiers, respectively. A comparison of these two t-SNE embeddings reveals that the hierarchical model more effectively groups child classes under the same parent class. This suggests that the hierarchical model captures the underlying taxonomy structure more accurately than the baseline flat model does. Additionally, the classes in the hierarchical model are more distinctly separated compared to those in the flat baseline model.

The improvement in performance of the hierarchical model, as evidenced by the AUROC scores, further substantiates that the hierarchically consistent loss function is effective in learning a more hierarchically consistent feature representation. This feature representation is subsequently used to calculate the Mahalanobis distance in DMD.

\section{Conclusion \label{sec:conclusion}}

We showed how, when available, hierarchical labels can be used to
improve novel fault class detection when deep neural network-based
fault classifiers are used. Integration of hierarchical information
can be intimidating, especially in the context of neural network training.
This is why we employ the most straightforward approach we could find
in the literature, the soft labels, as a proof of concept. We extended
the existing literature on novel class detection in deep learning
to its hierarchical counterpart.

Our claims are experimentally demonstrated on a real-life hot steel
rolling defect image dataset. Our experiment design emulates an environment
where a fault class has never been observed during training time and
is introduced at testing time. Accounting for variations in weight
initializations and hyperparameter optimizations, we demonstrate that
our approach increases the AUROC metric for all possible scenarios
and methodology extensions.

In this work, we are only focusing on utilizing the soft-label approaches
to encode hierarchical information as well as define the monitoring
statistics. There are some other hierarchical models and advanced
anomaly detection models, and how to propose proper test statistics
for those would be another interesting future work.

Additionally, we are also considering a number of open questions in
future work. Whether additional post-training calibration methods
can drive the detection performance up higher or not has been answered
in this work. Although we deliberately avoided the use of auxiliary
datasets in the context of fault classification because of its impracticality
in real-life scenarios, we suspect that simulated data may address
this issue. Another question we are interested in is whether a partially
unknown class (e.g., A12) can be detected along with the correct classification
at a coarse level. This is an important question for the practitioner
may choose to take different actions based on obliviousness to various
levels of the hierarchy.

\section{Data Availability Statement}

\label{sec:dataavail} Due to the proprietary nature of the collected
dataset from the industry, supporting data cannot be made openly available.
However, the code will be made available on the Github repo. Further
information about the codes will be made available at the Github repo
after the paper acceptance.

\bibliographystyle{chicago} \spacingset{1} \bibliographystyle{chicago}
\bibliography{IISE-Trans}

\clearpage
\pagenumbering{arabic}

\appendix

\section{Appendix}

\subsection{\label{subsec:Proof-of-Proposition 1} Proof of \Cref{Proposition-1: hiearchical-score} }
\begin{proof}
The hierarchical score is given as 
\[
\begin{aligned}s(f_{k}) & =-\sum_{k=1}^{K}l_{k}^{\text{soft}}(\hat{y})\log f_{k}(\mbx,\mbtheta^{*})\\
 & \text{Let }l(k)\text{ denote }l_{k}^{\text{soft}}(\hat{y})\text{, and }f(k)\text{ denote }f_{k}(\mbx,\mbtheta^{*})\\
 & =-\sum_{k}l(k)\log f(k)\\
 & =-\sum_{k}l(k)\log\frac{f(k)l(k)}{l(k)}\\
 & =-\sum_{k}l(k)\log\frac{f(k)}{l(k)}-\sum_{k}l(k)\ln{l(k)}\\
 & =D_{KL}(l\parallel f)+H(l)\geq H(L)\\
 & \text{The equality condition is satisfied if and only if }f(k)=l(k).
\end{aligned}
\]
where $D_{KL}(l\parallel f)$ denotes the KL divergence between $l$
and $f$, $H(l)$ denotes the entropy of $l(k)$, which is a constant
given the specified hierarchical structure and hyper-parameter $\beta$. 
\end{proof}

\subsection{\label{subsec:Proof-of-Proposition 2 }Proof of \Cref{prop2: temp-scaling}}

The softmax function with temperature scaling is given by

\[
\begin{aligned}f_{k}(\boldsymbol{x};T)&=\frac{\exp{(g_{k}(\mbx)/T})}{\sum_{j=1}^{K}\exp{(g_{j}(\mbx)/T)}}\\&=\frac{1}{1+\Sigma_{j\neq k}\exp({\frac{(g_j(\boldsymbol{x})-g_k(\boldsymbol{x)})}{T}})}
\\&\approx \frac{1}{1+(K-1)+\frac{1}{T}\Sigma_{j\neq k}^K (g_j(\boldsymbol{x})-g_k(\boldsymbol{x}))+o(\frac{1}{T^2})}
\\&\approx \frac{1}{K+\frac{1}{T}\Sigma_{j\neq k}^K (g_j(\boldsymbol{x})-g_k(\boldsymbol{x})))}
\end{aligned}
\]
\subsection{\label{subsec:Proof-of-Proposition 3}Proof of \Cref{prop3: perturbation} }

The perturbed version of the hierarchically consistent score is given
by
\begin{align*}
-s(f_{k}) & =\sum_{k=1}^{K}l_{k}^{\text{soft}}(\hat{y})\log f_{k}(\tilde{\mbx})\\
 & =l_{\hat{y}}^{\text{soft}}(\hat{y})\log f_{\hat{y}}(\tilde{\mbx})+\sum_{k\neq\hat{y}}l_{k}^{\text{soft}}(\hat{y})\log f_{k}(\tilde{\mbx})
\end{align*}

\begin{align*}
 & l_{\hat{y}}^{\text{soft}}(\hat{y})\log f_{\hat{y}}(\tilde{\mbx})\\
= & l_{\hat{y}}^{\text{soft}}(\hat{y})\log f_{\hat{y}}(\mbx)-\epsilon\text{sign}\left(-\nabla_{\mbx}\log f_{\hat{y}}(\mbx)\right)\cdot\left(\nabla_{\mbx}\log f_{\hat{y}}(\mbx)\right)\\
= & l_{\hat{y}}^{\text{soft}}(\hat{y})\log f_{\hat{y}}(\mbx)+\underbrace{\|\nabla_{\mbx}\log f_{\hat{y}}(\mbx)\|_{1}}_{-U_{1}}
\end{align*}

Similarly, 
\begin{align*}
 & \sum_{k\neq\hat{y}}l_{k}^{\text{soft}}(\hat{y})\log f_{k}(\tilde{\mbx})\\
= & \sum_{k\neq\hat{y}}l_{k}^{\text{soft}}(\hat{y})\log f_{k}(\mbx)-\sum_{k\neq\hat{y}}l_{k}^{\text{soft}}(\hat{y})\epsilon\text{sign}\left(-\nabla_{\mbx}\log f_{\hat{y}}(\mbx)\right)\cdot\left(\nabla_{\mbx}\log f_{k}(\mbx)\right)\\
= & \sum_{k\neq\hat{y}}l_{k}^{\text{soft}}(\hat{y})\log f_{k}(\mbx)\underbrace{-\sum_{k\neq\hat{y}}l_{k}^{\text{soft}}(\hat{y})\text{sign}(-\nabla_{\mbx}\log f_{\hat{y}}(x))\cdot\nabla_{\mbx}\log f_{k}(\boldsymbol{x})}_{-U_{2}}
\end{align*}

Also, it is easy to get the lower bound of $U_2$: 

$U_{2}=\sum_{k\neq\hat{y}}l_{k}^{\text{soft}}(\hat{y})\text{sign}(-\nabla_{\mbx}\log f_{\hat{y}}(x))\cdot\nabla_{\mbx}\log f_{k}(\boldsymbol{x})\geq -\sum_{k\neq\hat{y}}l_{k}^{\text{soft}}(\hat{y})\|\nabla_{\mbx}\log f_{k}(\boldsymbol{x})\|_{1}$. 

\subsection{\label{subsec:U1U2example} $U_1$ and $U_2$ values}

\begin{figure}[!t]
    \begin{subfigure}{0.5\linewidth}
        \centering
        \includegraphics[width=0.8\linewidth]{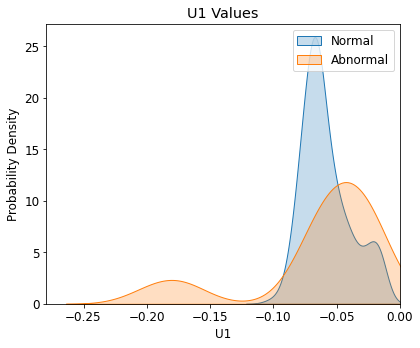}
        \caption{$U_1$ for normal and abnormal classes}
        \label{fig:U1}
    \end{subfigure}%
    \begin{subfigure}{0.5\linewidth}
        \centering
        \includegraphics[width=0.8\linewidth]{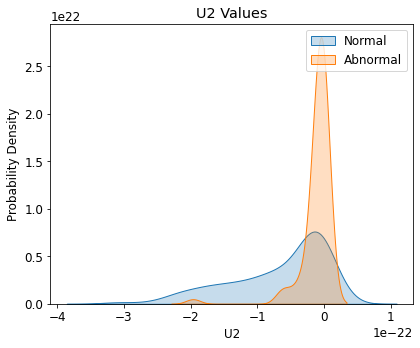}
        \caption{$U_2$ for normal and abnormal classes}
        \label{fig:U2}
    \end{subfigure}
    \caption{U1 and U2 values comparison}
    \label{fig:side_by_side}
\end{figure}

\Cref{fig:U1} and \Cref{fig:U2} present an illustrative example from the dataset in this study to elucidate  \Cref{rem: U1} and \Cref{rem: U2}. The normal class is more likely to exhibit smaller values than the abnormal classes for both $U_1$ and $U_2$ as discussed in \Cref{rem: U1} and \Cref{rem: U2}. This validates the assumptions that combining the hierarchical treatment and the perturbation can indeed separate the abnormal class from the normal class better.

\subsection{\label{subsec:Proof-of-Proposition 6}Proof of \Cref{prop: msp}}
\begin{proof}
$\hat{y}$ is defined as $\hat{y}=\argmax_{k\in\{1\dots K\}}\{f_{k}(\mbx,\mbtheta^{*})\}$.
The soft label embedding can be considered as the weights $l_{k}^{\text{soft}}(\hat{y})=\frac{\exp(-\beta d(k,\hat{y}))}{\sum_{j}^{K}\exp(-\beta d(j,\hat{y}))}.$
When $\beta\to+\infty$, we can get:

\[
\lim_{\beta\to\infty}l_{k}^{\text{soft}}(\hat{y})=\lim_{\beta\to\infty}\frac{\exp(-\beta d(k,\hat{y}))}{\sum_{j}^{K}\exp(-\beta d(j,\hat{y}))}=\left\{ \begin{aligned} & 1 &  & \text{if }k=\hat{y}\\
 & 0 &  & \text{if }k\neq\hat{y}
\end{aligned}
\right.
\]

\[
\begin{aligned} & \lim_{\beta\to\infty}-\sum_{k=1}^{K}l_{k}^{\text{soft}}(\hat{y})\log f_{k}(\mbx,\mbtheta^{*})\\
= & -\log f_{k=\hat{y})}(\mbx,\mbtheta^{*})=-\log\max f_{k}(\mbx,\mbtheta^{*})>c
\end{aligned}
.
\]
Then, the hierarchical score for novel fault detection can be equivalently
written as $\max f_{k}(\mbx,\mbtheta^{*})<exp(-c)=c'$. 
\end{proof}

\subsection{Sample Size Summary}

\label{app:samplesize} In our dataset used in the case
study, the sample sizes for classes vary from 18 to 135. The detailed
sample sizes for all classes are shown in \Cref{t:sample_size}.

\begin{table}[h]
\centering \caption{Sample size summary for all defects}
\begin{tabular}{|c|c|}
\hline 
Label  &  Sample Size \tabularnewline
\hline 
A10  &  44 \tabularnewline
\hline 
A11  & 92 \tabularnewline
\hline 
A12  & 75 \tabularnewline
\hline 
A20  &  135 \tabularnewline
\hline 
A21  & 54 \tabularnewline
\hline 
A30  &  127 \tabularnewline
\hline 
A31  &  115 \tabularnewline
\hline 
A40  &  18 \tabularnewline
\hline 
A41  & 105 \tabularnewline
\hline 
A50  &  89 \tabularnewline
\hline 
A51  & 78 \tabularnewline
\hline 
A60  & 72 \tabularnewline
\hline 
A61  &  75 \tabularnewline
\hline 
A70  & 96 \tabularnewline
\hline 
\end{tabular} \label{t:sample_size} 
\end{table}









\end{document}